\documentclass[10pt,twocolumn,letterpaper]{article}

\usepackage{cvpr}              

\usepackage[dvipsnames]{xcolor}

\definecolor{cvprblue}{rgb}{0.21,0.49,0.74}
\usepackage[pagebackref,breaklinks,colorlinks,citecolor=cvprblue]{hyperref}

\usepackage{graphicx}
\usepackage{amsmath}
\usepackage{amssymb}
\usepackage{booktabs}
\usepackage{subcaption}
\usepackage{float}
\usepackage{pifont}
\usepackage{multirow}
\usepackage[pagebackref,breaklinks,colorlinks]{hyperref}

\usepackage[capitalize]{cleveref}
\crefname{section}{Sec.}{Secs.}
\Crefname{section}{Section}{Sections}
\Crefname{table}{Table}{Tables}
\crefname{table}{Tab.}{Tabs.}

\newcommand{\smallsec}[1]{\vspace{0.2em}\noindent\textbf{#1}}

\def\paperID{931} 
\def\confName{CVPR}
\def\confYear{2024}

\title{Separate-and-Enhance: Compositional Finetuning for Text2Image Diffusion Models}

\newcommand{\modelname}[0]{SepEn\xspace} 
\newcommand{\modelnamestar}[0]{SepEn$^*$}

\author{
{Zhipeng Bao$^{1, 2}$ \qquad Yijun Li$^{2}$ \qquad Krishna Kumar Singh$^{2}$ \qquad Yu-Xiong Wang$^{3}$}  \qquad Martial Hebert$^1$ \\
{ $^1$CMU \qquad $^2$Adobe Research \qquad $^3$UIUC}\\
  \texttt{\footnotesize \{zbao, hebert\}@cs.cmu.edu \qquad \{yijli, krishsin\}@adobe.com \qquad yxw@illinois.edu} \\
}

\begin{document}

\twocolumn[{%
\renewcommand\twocolumn[1][]{#1}%
\maketitle
\begin{center}
    \includegraphics[width = \linewidth]{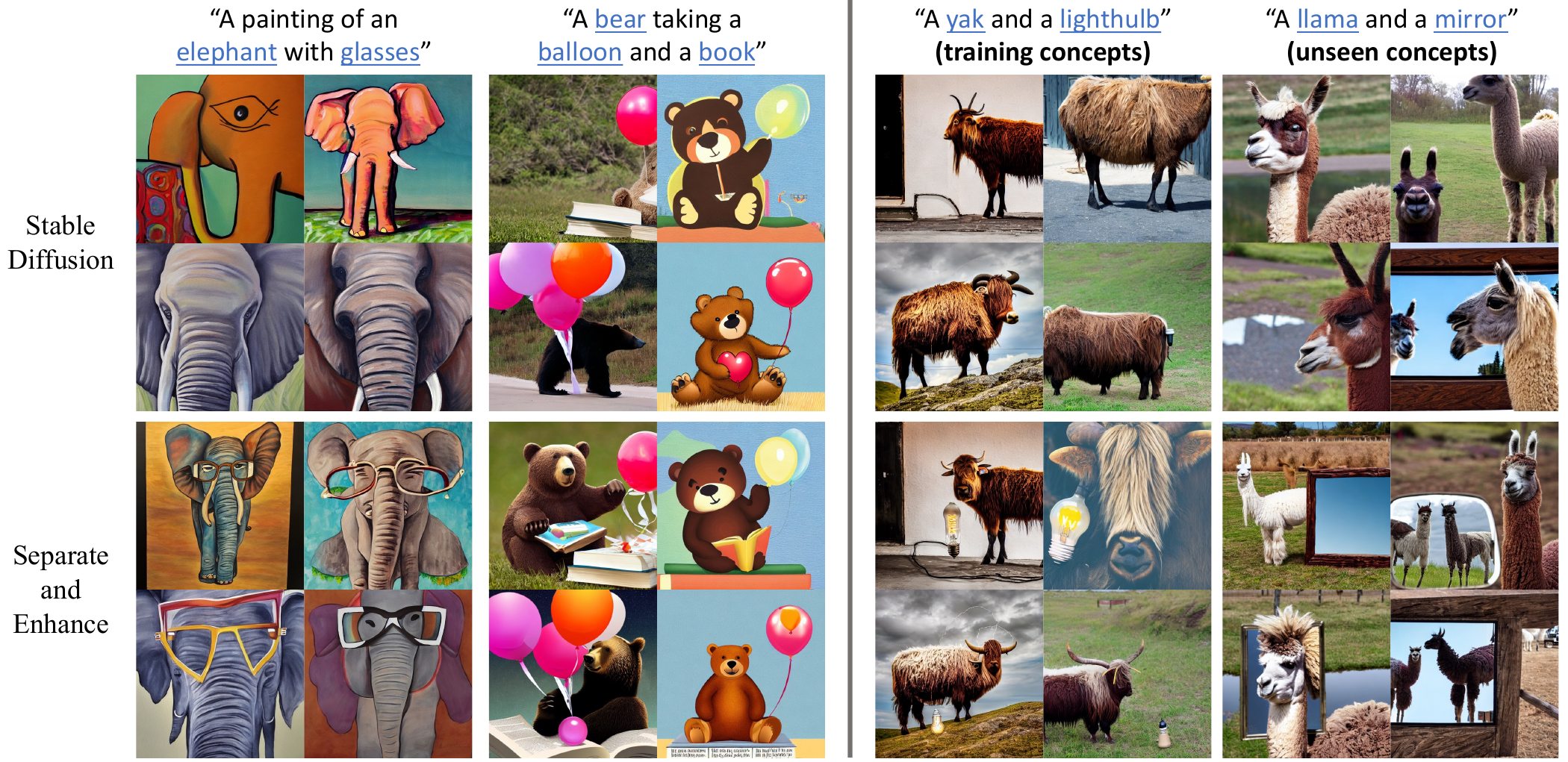}
    \vspace{-0.25in}
    \captionof{figure}{Visual comparisons between Stable Diffusion~\cite{rombach2022high} and our method. Left: compositional finetuning with individual concepts. After finetuning, we are able to generate more aligned images with the text input of high quality. Right: joint compositional finetuning with a large collection of concepts. After finetuning, the model keeps a high compositional capacity for unseen novel concepts.
    }
    \label{fig:cover}
\end{center}
}]

\begin{abstract}
    Despite recent significant strides achieved by diffusion-based Text-to-Image (T2I) models, current systems are still less capable of ensuring decent compositional generation aligned with text prompts, particularly for the multi-object generation. 
    This work illuminates the fundamental reasons for such misalignment, pinpointing issues related to low attention activation scores and mask overlaps. While previous research efforts have individually tackled these issues, we assert that a holistic approach is paramount. Thus, we propose two novel objectives, the Separate loss and the Enhance loss, that reduce object mask overlaps and maximize attention scores, respectively. 
    Our method diverges from conventional test-time-adaptation techniques, focusing on finetuning critical parameters, which enhances scalability and generalizability. Comprehensive evaluations demonstrate the superior performance of our model in terms of image realism, text-image alignment, and adaptability, notably outperforming prominent baselines. Ultimately, this research paves the way for T2I diffusion models with enhanced compositional capacities and broader applicability.
\end{abstract}

\section{Introduction}
\label{sec:introduction}

\begin{figure}
    \centering
    \includegraphics[width =  \linewidth]{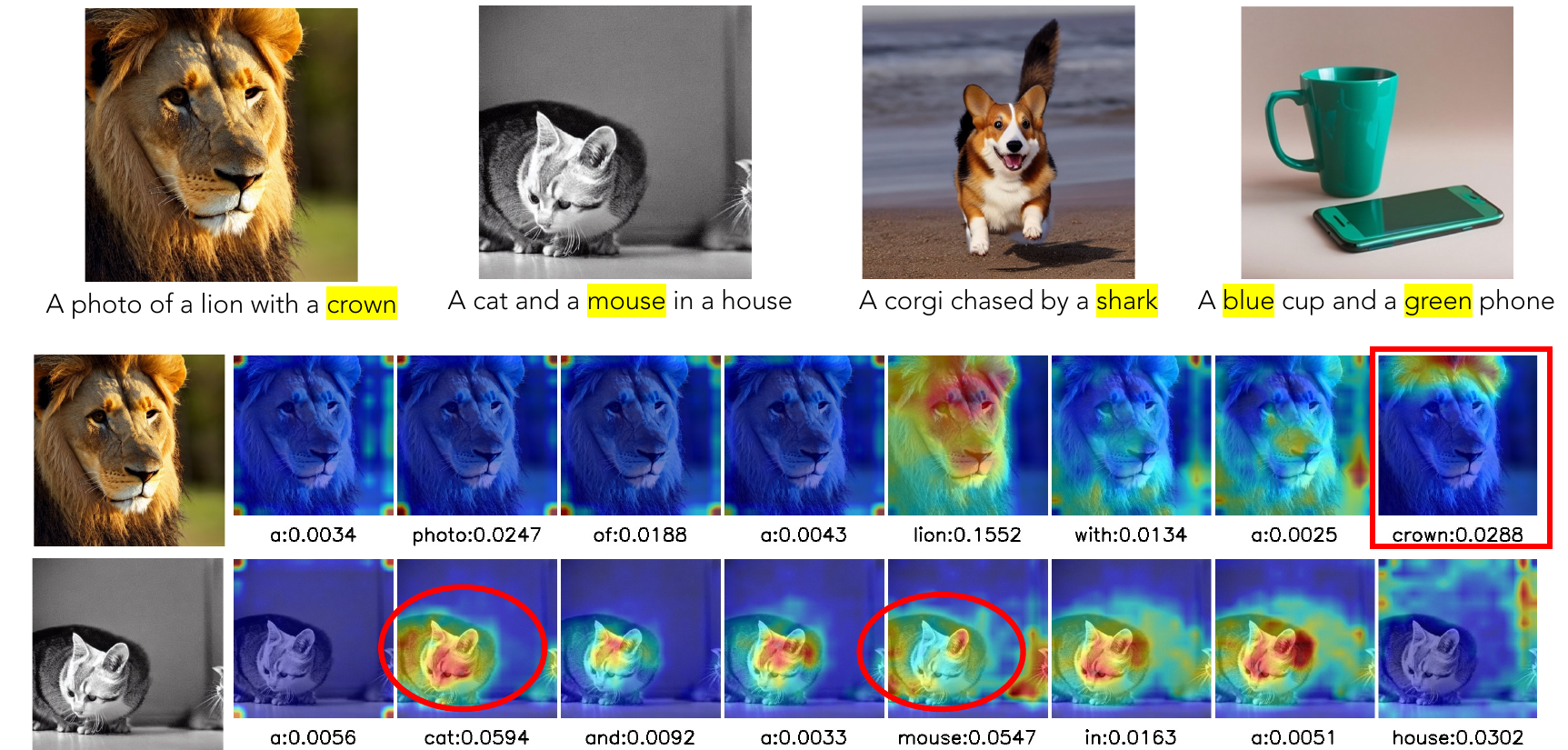}
    \caption{(Top) Failure cases and (Bottom) reasons of Stable Diffusion~\cite{rombach2022high}. Even state-of-the-art T2I models grapple with challenges when representing multiple objects with varying attributes. Two primary factors, demonstrated by the bottom two examples respectively, include: (1) low attention activation scores for certain objects and (2) the attention masks overlap.
    }
    \label{fig:teaser}
\end{figure}

Human cognition possesses a remarkable capacity for compositional understanding, allowing us to focus on, differentiate between, and even conceptualize novel objects within our environment~\cite{tomasello2009cultural}. Mirroring this capability within machine vision systems, especially in the realm of generative modeling such as Text-to-Image synthesis~\cite{ramesh2021zero,rombach2022high} which seeks to produce photo-realistic images that coherently represent given textual descriptions, has garnered significant attention in the computer vision community. However, even cutting-edge diffusion-based~\cite{ho2020denoising} T2I models, including the noted Stable Diffusion model by Rombach et al.\cite{rombach2022high}, grapple with challenges when representing multiple objects with varying attributes, such as different shapes, sizes, and colors as shown on top of  Figure~\ref{fig:teaser}. 
To understand the underlying reasons, we visualize the text-image correspondence for two examples with cross-attention masks at the bottom of Figure~\ref{fig:teaser}. Through our analysis, we find two primary factors for observed text-image misalignment: (1) the attention activation scores for certain objects are notably low and (2) the attention masks corresponding to diverse objects exhibit substantial overlap. Potential causes for these phenomena include the dominance of certain object classes among others, and the rarity of certain combinations during the model training. Fortunately, they can be addressed by tweaking the attention layers of the T2I models.

Previous research has identified similar reasons, and endeavored to address them~\cite{chefer2023attend,agarwal2023star,li2023divide,huang2023reversion}. Nonetheless, these approaches primarily concentrate on the singular facet of the problem: either amplifying attention activation~\cite{chefer2023attend,li2023divide} or reducing attention overlap~\cite{agarwal2023star,huang2023reversion}. Our empirical observations indicate that solutions focusing solely on one of these aspects yield limited enhancements to the compositionality of T2I models, as supported by our results in Table~\ref{tab:quantitative} and Figure~\ref{fig:large_scale}. Therefore, to better address these challenges, we propose two novel objectives respectively, the {\it Separate loss} designed to mitigate the Intersection of Union (IoU) of multiple objects and prevent them from coalescing into a singular entity; and the {\it Enhance loss}, which seeks to maximize the attention activation scores associated with each object. 

In light of the proposed objectives above, one question emerges: \emph{How to best execute compositional finetuning?} A considerable portion of previous work~\cite{chefer2023attend,agarwal2023star,li2023divide} employs a strategy akin to test-time adaptation. Specifically, they retain the weights of pre-trained T2I models and refine only the latent features for each pair of new concepts. Notable shortcomings of this type of approach include: (1) it fails to truly improve the compositional capacity of diffusion models as it remains training-free; (2) it results in longer inference times owing to the adaptations performed during testing; and most critically, (3) it is unable to scale up to {\it multiple} concepts which lack the {\bf generalizability} for new concepts. To address these limitations, we directly apply our objectives to finetune the diffusion model, aiming at enhancing the inherent compositional ability of pre-trained T2I models. Furthermore, a deep analysis of each function within the text-image attention modules enables us to selectively finetune a specific subset of parameters, the Key mapping functions (Section~\ref{sec:method}). This strategy leads to an overall lightweight finetuning process and makes it possible for our method to generalize to larger scales. 

To validate the effectiveness of our mechanism, we conduct a comprehensive experimental evaluation. Firstly, we evaluate our model with several individual prompt pairs. Our model could achieve a much higher success rate of text-image alignment and more realistic image generation compared with the baselines including the Stable Diffusion model~\cite{saharia2022photorealistic} and its following works~\cite{chefer2023attend,liu2022compositional} (see typical examples in Figure~\ref{fig:cover}, left). Sequentially, large-scale experiments on a collection 220 concepts from ImageNet-21K~\cite{deng2009imagenet} indicate that our method manages to simultaneously process multiple concept pairs. Notably, the resulting model from large-scale finetuning shows a great generalization capacity for unseen concepts  (refer to Figure~\ref{fig:cover}, right). Thirdly, we additionally ablate each of the two objectives to contribute to the final promising results. Lastly, we show that our finetuned model not only improves compositional generation of multiple objects, but also retains comparable performance for single-object synthesis. 

In conclusion, our contributions are threefold: {\bf (1)} We analyze the underlying factors responsible for the compositional misalignment in T2I diffusion models. Subsequent to this analysis, we introduce two novel objectives: the {\it Separate loss} to decouple the object masks and the {\it Enhance loss} to maximize the attention activation scores of each object; {\bf (2)} We carefully design a compositional finetuning scheme by only tweaking the most critical parameters, which ensures the scalability of our method across broader scales; {\bf (3)} Experimental evaluations validate that our algorithm can greatly improve the composition capacity of T2I diffusion models for both individual concept pairs and a large collection of concepts. Our model under the large-scale setting also exhibits great generalization to novel concepts.

\section{Related Work} 
\label{sec:related}

\smallsec{Text-to-Image Synthesis} aims to generate realistic and semantically consistent images from textual descriptions~\cite{reed2016generative}. A lot of previous efforts have made significant advancements in this space based on Generative adversarial networks(GANs)~\cite{goodfellow2014generative, xu2018attngan,zhang2017stackgan,bau2021paint,zhu2019dm}, Variational autoencoder (VAEs)~\cite{ramesh2021zero}, and diffusion models~\cite{rombach2022high,nichol2021glide,saharia2022photorealistic,ramesh2022hierarchical,gu2022vector,yu2022scaling}. Among them, 
diffusion models have emerged as the most promising approach and showed the state-of-the-art generation quality. These models are built upon principles of heat diffusion and anisotropic diffusion to process and generate images~\cite{perona1990scale,weickert1998anisotropic}. Combining diffusion and neural networks results in powerful generative models such as the Denoising Diffusion Probabilistic Model (DDPM)~\cite{ho2020denoising} and the Score-Based Generative Model (SBGM)~\cite{song2020score}. 
The text information is usually introduced via the cross-attention~\cite{vaswani2017attention} unions during the denoising process. These methods showcase the adaptability and effectiveness of diffusion models in various text-based synthesis and editing applications.

\smallsec{Compositional synthesis with diffusion models.} 
Existing T2I models have been observed to be less capable of generating multiple objects described in prompts which essentially indicates their poor compositional ability and thus causes text-image misalignment. To solve this issue, there are two main branches of the methods to enhance the compositional text-image alignment for T2I models. The first branch of the work tackle the task in a style of test-time-adaptation by running attention guidance~\cite{hertz2022prompt,hong2022improving,wang2023compositional,epstein2023diffusion}, tweaking attention masks~\cite{chefer2023attend,li2023divide,agarwal2023star} or refining latent representations~\cite{liu2022compositional,brooks2023instructpix2pix}. Among them, Attend-and-Excite~\cite{chefer2023attend} encourages the attention activation for all the objects to be as strong as possible. A most recent work, A-Star~\cite{agarwal2023star}, designs novel loss functions to split the objects from one another. Though these two works share a similar spirit of our method, one major limitation is that they are handling concepts pair by pair which lacks the {\it generalizability} to novel concepts. Meanwhile, they are inference time solutions that increase runtime and require parameter tuning per pair. 
The other type of methods further finetune the pre-trained diffusion model with additional structured inputs or supervisions, such as masks~\cite{wang2022semantic,zhang2023adding,huang2023reversion}, bounding boxes~\cite{zheng2023layoutdiffusion,ma2023directed}, and external vision-language models~\cite{feng2022training,singh2023divide}. While more signals beyond text will definitely help reduce the text-image misalignment issue, our work is still focusing on improving the compositional ability of the T2I model itself given the text prompt only. We \emph{do not} require any additional supervision during both training and inference time. Our goal is that our finetuned model not only improves the compositional generation of multiple objects, but also keeps comparable performance for single-object synthesis, so that eventually only one single model is needed for direct inference from just the text prompt input. 

\section{Proposed Method}
\label{sec:method}

\subsection{Preliminary}

\smallsec{Stable Diffusion.} We build our model upon the state-of-the-art Stable Diffusion (SD)~\cite{rombach2022high}. Different from the basic pixel-based diffusion models~\cite{ho2020denoising,song2020denoising}, SD runs in the latent space of the autoencoder rather than the image space. First, an encoder $\mathcal{E}$ is trained to map a given image $x \in \mathcal{X}$ into a spatial latent code $z =
\mathcal{E}(x)$. A decoder $\mathcal{D}$ is then tasked with reconstructing the input image such that $\mathcal{D}(\mathcal{E}(x)) \approx x$.

SD first pre-trains the large-scale autoencoder, then they further train a DDPM model that operates over the learned latent space to produce a denoised version of a noise input $z_t$ at time step $t$. During the denoising process, the diffusion model can be conditioned on an additional input vector such as the text, layout, or semantic map~\cite{rombach2022high}. Concretely, in SD, the conditional input is a prompt embedding produced by a pre-trained CLIP text encoder~\cite{radford2021learning}. Given a conditional embedding $c(y)$, which conditioned on the prompt $y$, the DDPM model $\epsilon_\theta$ is trained to minimize the loss,
\begin{equation}
    \mathcal{L} = \mathbb{E}_{z \sim \mathcal{E}(x), y,\epsilon \sim \mathcal{N}(0,1),t} || \epsilon - \epsilon_\theta(z_t, t, c(y))||^2.
    \label{eq:sd}
\end{equation}
In summary, at each time step $t$, $\epsilon_\theta$ is tasked with correctly removing the noise $\epsilon$ added to the latent code $z$, given the noised latent $z_t$, timestep $t$, and conditioning encoding $c(y)$. $\epsilon_\theta$ is a network with UNet architecture~\cite{ronneberger2015u} consisting of self-attention and cross-attention layers, as discussed below.

\smallsec{Text Condition.} SD uses text embedding to guide the image thesis through cross-attention modules. Their cross-attention modules contain a self-attention layer followed by a cross-attention layer in each group. They run on resolutions 64, 32, 16, and 8. The text embedding is added to guide the image generation via the cross-attention layers.

Concretely, the latent vector input of the $i^{th}$ layer of the UNet, $z^{i}_t$, works as the query of the cross-attention, the projections of the text embedding $c(y) \in \mathbb{R}^{N \times d_\mathrm{text}}$, where $N$ is the length of the text sequence and $d_\mathrm{text}$ is the embedding dimension, is used as the key and value of the multi-head attention (MHA)~\cite{vaswani2017attention} module:
\begin{equation}
    z^{i'}_t = \text{MHA}(q(z^i_t),~k(c(y)),~v(c(y))).
\end{equation}
By running the attention operation, we can also get the attention masks $M_t \in \mathbb{R}^{N \times (p \times p)}$ for further processing, where $p \in \{ 64,32,16,8\}$ is the spatial resolution of the latent feature. For our method, we operate on attention masks under resolution $16 \times 16$. For all the following texts, we note $M_t \in \mathbb{R}^{N \times (16 \times 16)}$ as the averaged attention masks for all the cross attention modules with resolution $16 \times 16$.

\begin{figure}
    \centering
    \includegraphics[width = \linewidth]{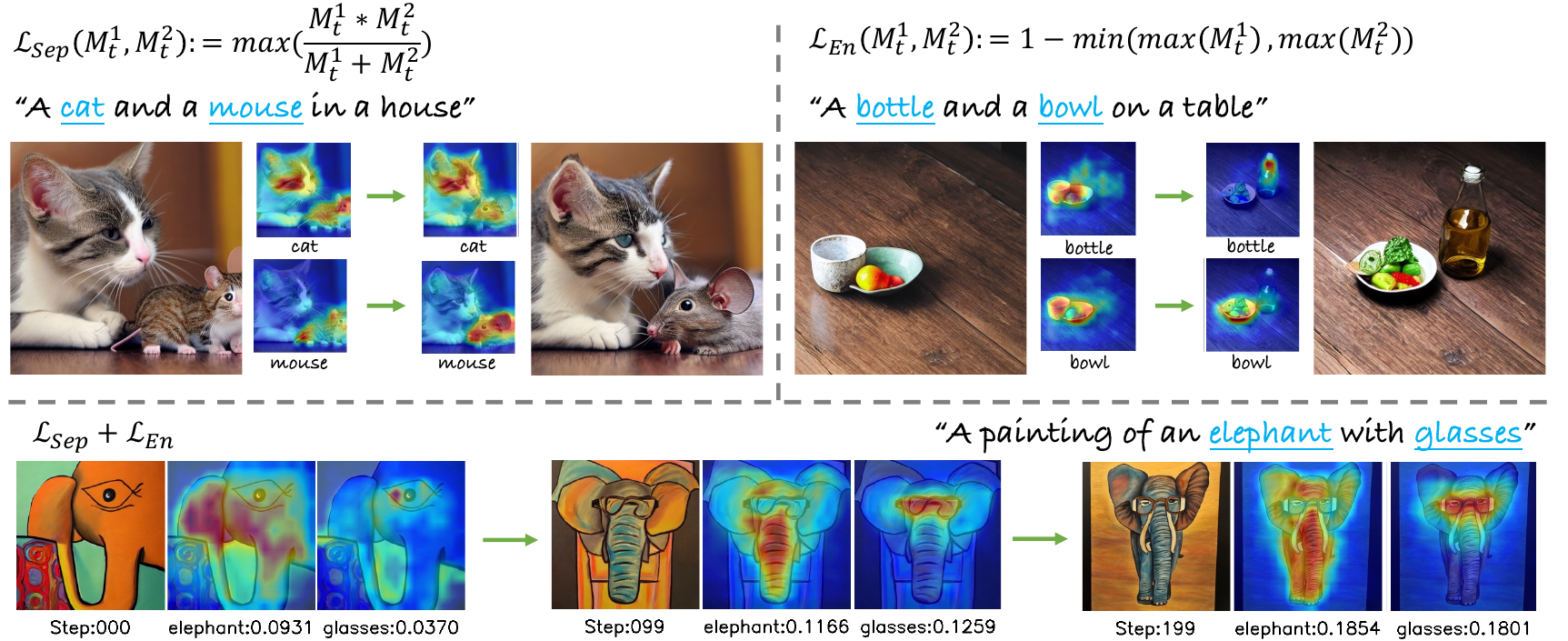}
    \vspace{-0.2in}
    \caption{Example of before and after applying our separate loss $\mathcal{L}_\mathrm{Sep}$. Left: attention maps from SD. Note that the \emph{bottle} attention falls onto the \emph{bowl} which results in failing to generate the bottle. Right: attention maps from our model where the attention of \emph{bottle} and \emph{bowl} are better separated so that each concept is validly generated.
    }
    \label{fig:seploss}
\end{figure}

\subsection{Separate-and-Enhance Pipeline}

As illustrated in Figure~\ref{fig:teaser}, two essential problems for the compositional misalignment are attention mask overlaps and low activation scores. We propose the Separate loss and the Enhance loss to tackle these two problems respectively as follows.

\smallsec{The Separate Loss.}
As discussed in Section~\ref{sec:introduction}, one reason for the compositional misalignment of the T2I model is the attention masks overlapping. As further shown in Figure~\ref{fig:seploss} (left), the original Stable Diffusion model fails to generate the \emph{bottle} in the given prompt. By visualizing the attention masks, we find that the primary reason is the attention of the \emph{bottle} falls onto the ``bowl'' region in the image. 

Therefore, we propose the Separate loss which aims at splitting object masks from binding to the same object region with the hope that salient objects should have as few overlapping as possible (refer to Figure~\ref{fig:seploss}, right). Concretely, during training, we randomly sample a timestep $t$ and obtain the attention masks for all the $K$ objects $\{M_t^i\}_{i=1}^K$. The Separate loss minimizes the maximum interaction of union for all the pixels as follows:
\begin{equation}
    \mathcal{L}_\mathrm{Sep} = max(\frac{\prod_{i=1}^K M^i_t}{\sum_{i=1}^K M^i_t}),
    \label{eq:Sep_loss}
\end{equation}
where $\prod$ indicates pixel-level product. 

\begin{figure}
    \centering
    \includegraphics[width = \linewidth]{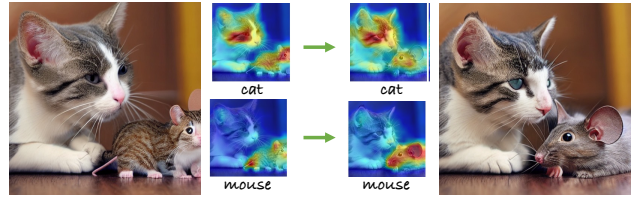}
    \vspace{-0.2in}
    \caption{Example of before and after applying our enhance loss $\mathcal{L}_\mathrm{En}$. Left: attention maps from SD. The attention score of \emph{mouse} is lower than that of \emph{cat} on the bottom right region, resulting in generating a cat-like mouse. Right: attention maps from our model where the attention activation score of \emph{mouse} is enhanced so that it can be correctly generated.
    }
    \label{fig:enloss}
\end{figure}

\smallsec{The Enhance Loss.}
Another reason for the text-image misalignment is the low activation for certain objects. From Figure~\ref{fig:enloss} (left), we see that the attention score of \emph{mouse} is lower than that of \emph{cat} on the bottom right region, resulting in generating a cat-like mouse. To solve this problem, we propose the Enhance loss aiming to highlight the attention activation score for all the objects with the hope that the synthesized images should have saliency regions for all the objects from the input prompts (refer to Figure~\ref{fig:enloss}, right).

Before computing the loss, we first inspired by~\cite{chefer2023attend}, we first filter the attention masks with a Gaussian smooth kernel~\cite{babaud1986uniqueness} to obtain the smoothed version of the masks $\{\tilde{M}_t^i\}_{i=1}^K$. Next, the Enhance loss will amplify the attention of the lowest-scored concept by minimizing
\begin{equation}
    \mathcal{L}_\mathrm{En} = 1 - min(max(\tilde{M}_t^1),\cdots, max(\tilde{M}_t^K)).
    \label{eq:En_loss}
\end{equation}

\begin{figure}[t]
    \centering
    \includegraphics[width = \linewidth]{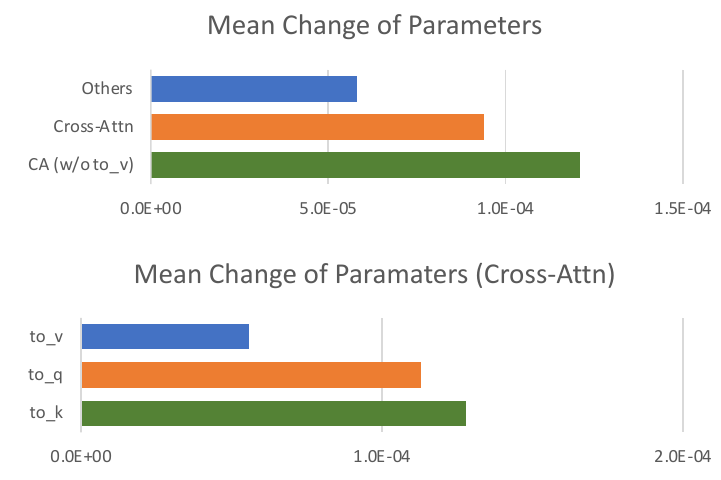}
    \caption{Average parameter changes for the whole network (top) and inside cross-attention modules (bottom) during finetuning. The parameters in the cross-attention modules are more sensitive to finetuning, especially for the key mapping functions.
    }
    \label{fig:delta}
\end{figure}

\begin{figure*}
    \centering
    \includegraphics[width = \linewidth]{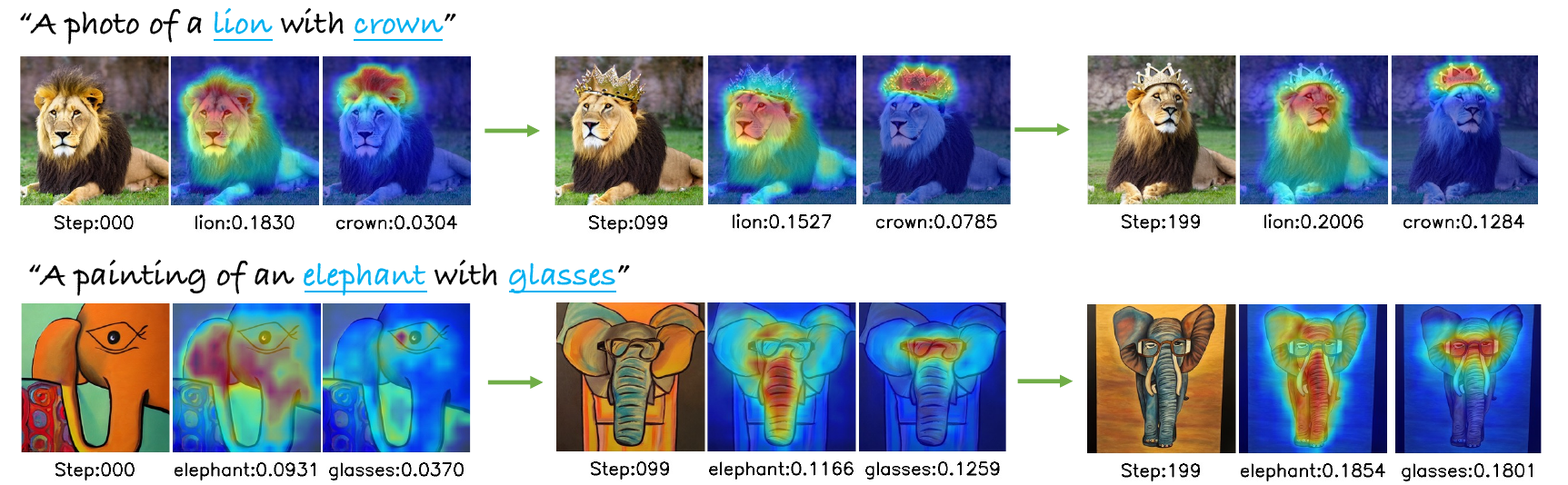}
    \caption{Example of change of generated images and object attention during different finetuning steps. The attention of two concepts in each row are gradually splitted and amplified until forming the final aligned image with high quality, showing the effectiveness of our proposed \modelname algorithm. 
    }
    \label{fig:bothloss-iteration}
    \vspace{-0.1in}
\end{figure*}

\subsection{Optimization}
\smallsec{Tuned parameters.} To efficiently and effectively do the compositional finetuning, inspired by Custom Diffusion~\cite{kumari2022multi}, we do a pilot experiment by finetuning the whole UNet architecture with 20 pairs of concepts. The average parameter changes for different modules are shown in Figure~\ref{fig:delta}. We find that the parameters in the cross-attention module are more sensitive to finetuning (top row of  Figure~\ref{fig:delta}) compared with the other modules. Therefore, we decided to only finetune the parameters in the cross-attention unions. Moreover, we discover that not all the parameters in the cross-attention unions should be tuned. We first show the general pipeline of the attention module with query (latent feature $z$ and mapping function $Q$), key (prompt $e_t$ and mapping function $K$), and value ($e_t$ and mapping function $V$) as follows:
\begin{equation}
    M = Q(z) K(e_t); ~~~~ z_\mathrm{out} = \mathrm{softmax}(M) V(e_t).
    \label{eq:qkv}
\end{equation}
It is noted from Equation~\ref{eq:qkv} that the $Q$ function maps the input noisy latent, which has few connections to the misalignment issue; and the $V$ function represents the feature embedding learned by the stable diffusion model to form the final latent vector for VAE synthesis, which we also don't want to finetune. Therefore, we select only the key mapping function $K$ as the target parameter group to finetune, leading to a lightweight overall finetuning strategy. This conclusion is also consistent with the results of Figure~\ref{fig:delta} (bottom) where the Key mapping function is the most sensitive one inside the cross-attention modules while the value mapping functions are the least. We also showcase in Section~\ref{sec:ablation} that only finetuning the $K$ function is the optimal choice compared with other variants.

\smallsec{Normalization term.}
As the proposed loss functions may lead to a distribution shift of the pre-trained SD model, especially for large-scale finetuning, we also add a standard normalization term similar to Equation~\ref{eq:sd}:
\begin{equation}
    \mathcal{L}_\mathrm{norm} = \sum_i \mathbb{E}_{z^i \sim \mathcal{E}(x^i), y^i,\epsilon \sim \mathcal{N}(0,1),t} || \epsilon - \epsilon_\theta(z^i_t, t, c(y^i))||^2,
    \label{eq:norm}
\end{equation}
where $(x^i, y^i)$ is the image-prompt pair we sampled with frozen pre-trained stable diffusion. The final loss objective is:
\begin{equation}
    \mathcal{L}_\mathrm{final} = \lambda_\mathrm{n} \mathcal{L}_\mathrm{norm} + \lambda_\mathrm{D} \mathcal{L}_\mathrm{En} + \lambda_\mathrm{E} \mathcal{L}_\mathrm{Sep},
\end{equation}
where $\lambda_\mathrm{n}, \lambda_\mathrm{E}$, and $\lambda_\mathrm{D}$ are weight factors. 

\smallsec{Synergy of the two objectives.} Notice that the two objectives tackle different areas of text-image misalignment while they are not totally disentangled: splitting objects from overlapping with each other helps to better enhance the low-activated ones, while making sure all the objects have high activation will also help to better detect the overlap in reverse. The synergy of the two objectives jointly improves the compositional capacity of T2I models. In Figure~\ref{fig:bothloss-iteration}, we show the change of images and object attention during different finetuning steps. The attention of both \emph{elephant} and \emph{glasses} are gradually splitted and amplified until forming the final aligned image with high quality, showing the effectiveness of our proposed \modelname algorithm.

\section{Experimental Results}
\label{sec:experiments}

\smallsec{Baselines.} We compared our model with three other diffusion-based state-of-the-art text2image models. \textbf{SD}~\cite{rombach2022high} is a powerful T2I model trained on large scale LAION-5B~\cite{schuhmann2022laion} with billions of annotated pairs. The other baselines and our method all build upon this pre-trained model.  \textbf{Attend-and-Excite (A \& E)}~\cite{chefer2023attend} aims to maximize the attention value of the target nouns during the inference time while \textbf{A-Star}~\cite{agarwal2023star} mitigates the attention mask overlaps. We report the original scores of A-Star due to a lack of open-source implementation. We also additionally report the results of Composable Diffusion~\cite{liu2022compositional} and Structure Diffusion~\cite{feng2022training} on single-prompt evaluations for reference.

\begin{figure*}[t]
    \centering
    \includegraphics[width = \linewidth]{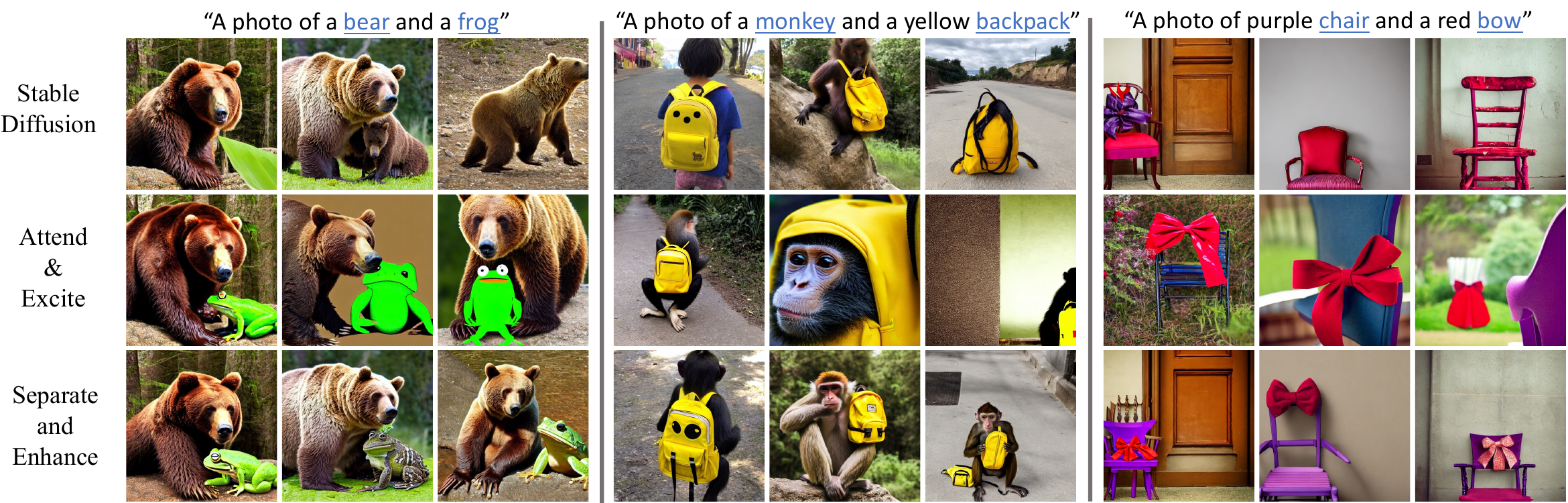}
    \vspace{-0.2in}
    \caption{Qualitative comparisons for our model and baselines. The images synthesized by our method have the best compositional alignment compared with the other two baselines. Meanwhile, our method also maintains a good visual quality for the generated images compared with the A \& E baseline~\cite{chefer2023attend}. 
    }
    \label{fig:qualitative}
    \vspace{-0.1in}
\end{figure*}

\smallsec{Eveluation prompts.} All of the compared methods do not require additional annotated data for finetuning thereby no datasets are used for our experiments. However, we do design two groups of evaluation protocols: individual prompts evaluation and large-scale evaluation. For the individual prompts evaluation, we use the same group of test prompts from Chefer \etal~\cite{chefer2023attend} that contains 276 prompts in three types: animal-animal, animal-object, and object-object. For the large-scale evaluation, we follow the process of \cite{chefer2023attend} and select 220 concepts (110 animals and 110 objects) from ImageNet-21~\cite{deng2009imagenet} categories. We randomly select 10 animals and 10 objects as held-out categories.

\smallsec{Metrics.} For the quantitative measurement, we adopt three metrics: (1) The {\bf FID score} measures the realism of the generated images, by computing the Fr{\'e}chet distance between two Gaussians fitted to feature representations of the source images and the target images~\cite{dowson1982frechet,parmar2021cleanfid}. Given that SD is renowned for its capacity to generate high-quality images, and one of our objectives is to preserve this image fidelity post-finetuning, we use generated images for \emph{single-object} prompts from SD as the source images. Notice that FID is not applicable to the Stable Diffusion baseline, mainly due to its tendency to underperform in generating images with multiple objects, despite generating photorealistic single-object images. However, we still report its FID for single-concept evaluation in Table~\ref{tab:quantitative} for reference; (2) the {\bf Average text-text Similarity Score} proposed by~\cite{chefer2023attend} measures if the generated contents match the input prompts. For each prompt, we compute the average BLIP~\cite{li2022blip} cosine similarity between the text prompt and the corresponding set of generated images; (3) the {\bf Success Rate} measures if the output images contain all objects mentioned in the text prompt. Specifically, we use a pre-trained detection model~\cite{zhou2022detecting} on ImageNet-21K to detect all possible objects from the given prompt.  We count as a success case if the highest confidence scores for all target objects are larger than 0.7.

\begin{table}[t]
\centering
\resizebox{\linewidth}{!}{
\begin{tabular}{l|ccc} 
\toprule
 Method & FID ($\downarrow$) & Average Similarity Score ($\uparrow$) & Success Rate ($\uparrow$) \\
 \hline
 StableDiffusion & \color{gray} 32.96 &  0.742$\pm$0.091 & 0.209 \\ 
 + Composable~\cite{liu2022compositional} & - &\color{gray} 0.71 & - \\
 + Structure~\cite{feng2022training} & - & \color{gray} 0.77 & - \\
 + A \& E~\cite{chefer2023attend} & 45.65 & 0.793$\pm$0.088 & 0.383 \\
 + A-Star~\cite{agarwal2023star} & - & \color{gray}  0.83 &  -\\
 + SepEn & \bf 36.85 & 0.809 $\pm$ 0.086 & \bf 0.410 \\ \hline
 + SepEn (TTA) & 41.74 &\bf 0.834 $\pm$ 0.081 & \bf 0.441 \\
\bottomrule
\end{tabular}
}
\caption{Quantitative comparisons for our model and baselines. Gray values indicate publicly reported numbers or for reference. Compared with the baselines, our method has a better compositional alignment with the best Average Similarity Score and Success Rate. Our \modelname also has a lower FID compared with A \& E ~\cite{chefer2023attend}, showing the good visual quality of our generated images. With the test-time adaptation, we are able to further improve the performance. }
\label{tab:quantitative}
\vspace{-0.2in}
\end{table}

\smallsec{Implementation Details.} We use SD 1.4 as the pre-trained model for a fair comparison with previous works~\cite{chefer2023attend,agarwal2023star}. We set $\lambda_E$ to 1.0 and $\lambda_E$ to 2.0 based on experience and attempts. We set $\lambda_N$ to 0.0 for individual training and 0.5 for large-scale training. We tune our method for 200 steps for each pair of individual concepts with a batch size of 4 on a single NVIDIA A-100 GPU. For the large-scale experiments, we tuned our model for 10,000 steps. We use the official implementation of A \& E with default configurations, and reproduce the A-Star method ourselves on the codebase of A \& E. More details are included in the appendix. 

\begin{figure}[t]
    \centering
    \includegraphics[width = \linewidth]{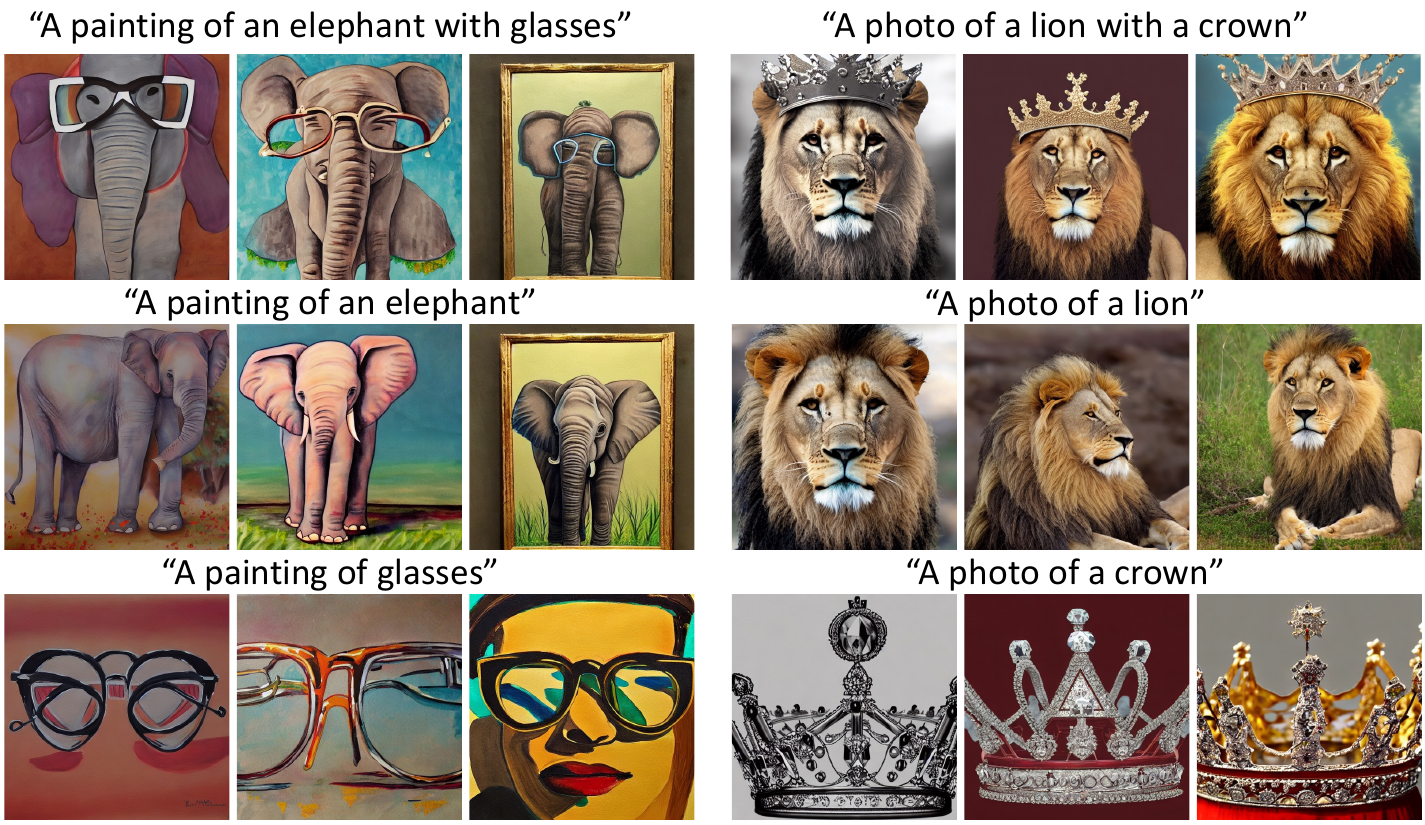}
    
    \caption{Single object synthesis with our tuned model. The resulting model still maintains a similar level of single-object synthesis as the original Stable Diffusion model.}
    \label{fig:single}
    \vspace{-0.1in}
\end{figure}

\subsection{Individual Prompts Evaluation}

\begin{figure*}[t]
    \centering
    \includegraphics[width = \linewidth]{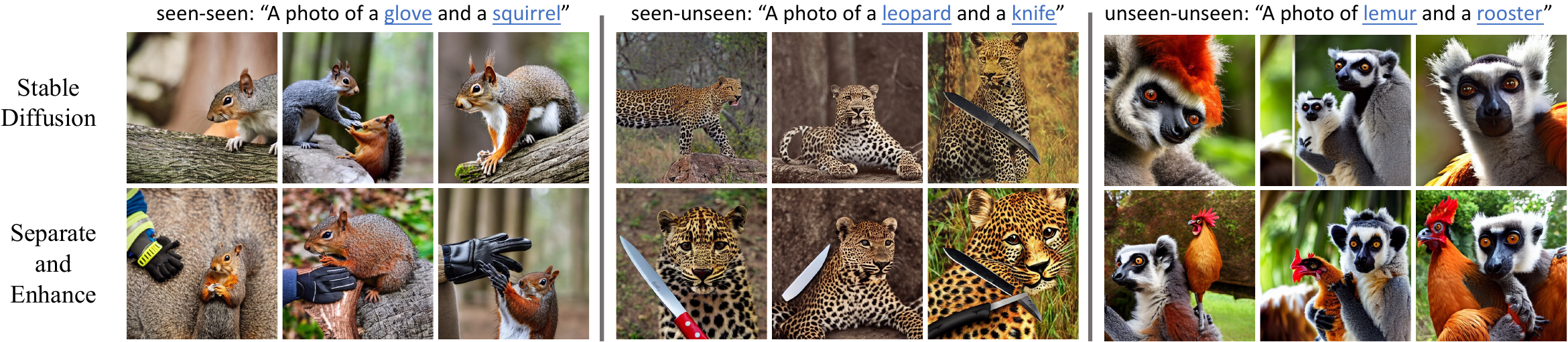}
    \caption{Visualizations for the large-scale experiments. Our model generates better text-aligned images with high quality for all three evaluation settings, indicating that our \modelname method has the capacity to jointly optimize a large collection of concepts with great generalizability to unseen novel concepts.}
    \label{fig:large_scale}
    \vspace{-0.1in}
\end{figure*}

\begin{table*}[t]
\centering
\resizebox{\linewidth}{!}{
\begin{tabular}{l|cc|cc|cc} 
\toprule
\multirow{2}{*}
{\shortstack{Method}} & \multicolumn{2}{c|}{seen-seen}  & \multicolumn{2}{c|}{seen-unseen}  & \multicolumn{2}{c}{unseen-unseen} \\ \cline{2-7}
    & Average Sim. Score ($\uparrow$) & Success Rate ($\uparrow$) & Average Sim. Score ($\uparrow$) & Success Rate ($\uparrow$) & Average Sim. Score ($\uparrow$) & Success Rate ($\uparrow$) \\
 \hline
 StableDiffusion  & 0.641 $\pm 0.107$ &0.212& 0.640 $\pm$ 0.105 & 0.227 & 0.633 $\pm$ 0.098 & 0.203\\ 
 + SepEn  &\bf 0.686 $\pm$ 0.107 &\bf 0.299 &\bf 0.677 $\pm$ 0.111 & \bf 0.305 &\bf 0.679 $\pm$ 0.102 & \bf 0.294 \\
 \bottomrule
\end{tabular}
}
\caption{Quantitative results for our method and the SD baseline. Ours significantly outperforms SD under all three regimes, indicating the great scalability of our model to a large collection of concepts and promising generalizability to novel concepts.}
\label{tab:large}
\vspace{-0.1in}
\end{table*}

\smallsec{Additional setup.} As the two baselines~\cite{chefer2023attend,agarwal2023star} run in a different manner from ours and they refine the latent outputs via test-time adaptation (TTA), we additionally report the quantitative results of a variant of our method, {\it SepEn (TTA)}, which follows the same test-time adaptation manner as the two baselines. It applies the two proposed objectives to refine the latent at each time step during inference and can be referred to as the performance upper bound for our model. 

\smallsec{Comparisons with baselines.} We show the qualitative and quantitative comparisons with baselines in Figure~\ref{fig:qualitative} and Table~\ref{tab:quantitative} respectively. More visualizations are included in the appendix. From those results, we observed that: (1) stable diffusion usually generates photo-realistic while misaligned single-object images, demonstrating the common issue of compositional capacity. Compared with SD, our approach is able to generate images significantly better aligned with the input prompts while keeping a good realism at the same time; (2) for the A \& E baseline, though it can generate images better aligned with the input prompts, the visual quality dropped a lot. We outperform this model under both realism and text-image alignment, owing to the two proposed novel objectives; (3) Even under the finetuning setting, we already outperform most of the TTA-based baseline models under all the metrics. With the test-time adaptation setting, our model outperforms all the baselines, including the most recent A-Star, for the text-image alignment. However, the visual quality dropped compared with our finetuned model. 

\smallsec{Single object generation.} In Figure~\ref{fig:single}, we show the synthesized images for the prompt with two objects and every single object. The finetuned model still maintains a similar level of single-object synthesis as the original SD model, indicating the robustness of our model's finetuning process.

\begin{table}[t]
\centering
\resizebox{\linewidth}{!}{
\begin{tabular}{l|ccc} 
\toprule
 Method & FID ($\downarrow$) & Average Similarity Score ($\uparrow$) & Success Rate ($\uparrow$) \\
 \hline
 StableDiffusion & \color{gray} 32.96 &  0.742$\pm$0.091 & 0.209 \\ 
 + $\mathcal{L}_\mathrm{Sep}$ & \bf 36.33 & 0.761$\pm$0.080 & 0.363 \\
 + $\mathcal{L}_\mathrm{En}$  & 42.84 & 0.770$\pm$0.093 &  0.374\\ \hline 
 + SepEn &  36.85 & \bf 0.809 $\pm$ 0.086 & \bf 0.410 \\
 \bottomrule
\end{tabular}
}
\caption{Ablation study with the Separate loss and the Enhance loss. The Enhance loss works better to improve the text-image compositional alignment while the Separate loss works better to keep the realism of the generated images.
Balancing them yields state-of-the-art performance.  }
\label{tab:loss}
\vspace{-0.2in}
\end{table}

\subsection{Large-scale Experiments}

For our model trained in large-scale, we propose three different types of evaluations: (1) \textbf{seen-seen} evaluation contains 100 pairs of concepts randomly sampled from the 200 training concepts; (2) \textbf{seen-unseen} evaluation contains 80 pairs of concepts, for each of the concepts in the held-out category, we randomly select one concept in the training set; (3) \textbf{unseen-unseen} evaluation contains 20 pairs of concepts, we randomly sample another concept from the held-out set for each unseen concept. We only compare with the Stable Diffusion baseline in this section, as the other baselines cannot generalize to a large scale of concepts. We report the Average Similarity Score and Success Rate. We also set the confidence threshold of the detector to 0.3 as some of the concepts are even very challenging for the detector.

The qualitative and quantitative results are reported in Figure~\ref{fig:large_scale} and Table~\ref{tab:large} respectively. 
Firstly, our model significantly outperforms the Stable Diffusion baseline both qualitatively and quantitatively, indicating the decent scalability of our model to a large collection of concepts, owing to our compositional finetuning strategy. Next, our model generalizes well to \emph{novel} (unseen) concepts with similar BLIP similarity score and success rate as the training ones, even for the most challenging unseen-unseen pairs, though the visual quality is a bit inferior to those pairs of trained (seen) concepts.
This result further convinces our goal of \emph{general compositional finetuning}. Finally, we find that there is still a large room for improvement regarding the overall success rate, indicating the general challenge of the text-image alignment task of T2I models.

\subsection{Ablation Study}
\label{sec:ablation}

\smallsec{Contributions of the two objectives.} We ablate the contributions of the Separate loss and the Enhance loss in Table~\ref{tab:loss}. We find that the Enhance loss works better to improve the text-image compositional alignment while the Separate loss works better to keep the realism of the generated images. Both objectives can bring improvements to the compositional capacity of diffusion T2I models, and balancing them yields state-of-the-art performance.

\begin{figure}[t]
    \centering
    \includegraphics[width = \linewidth]{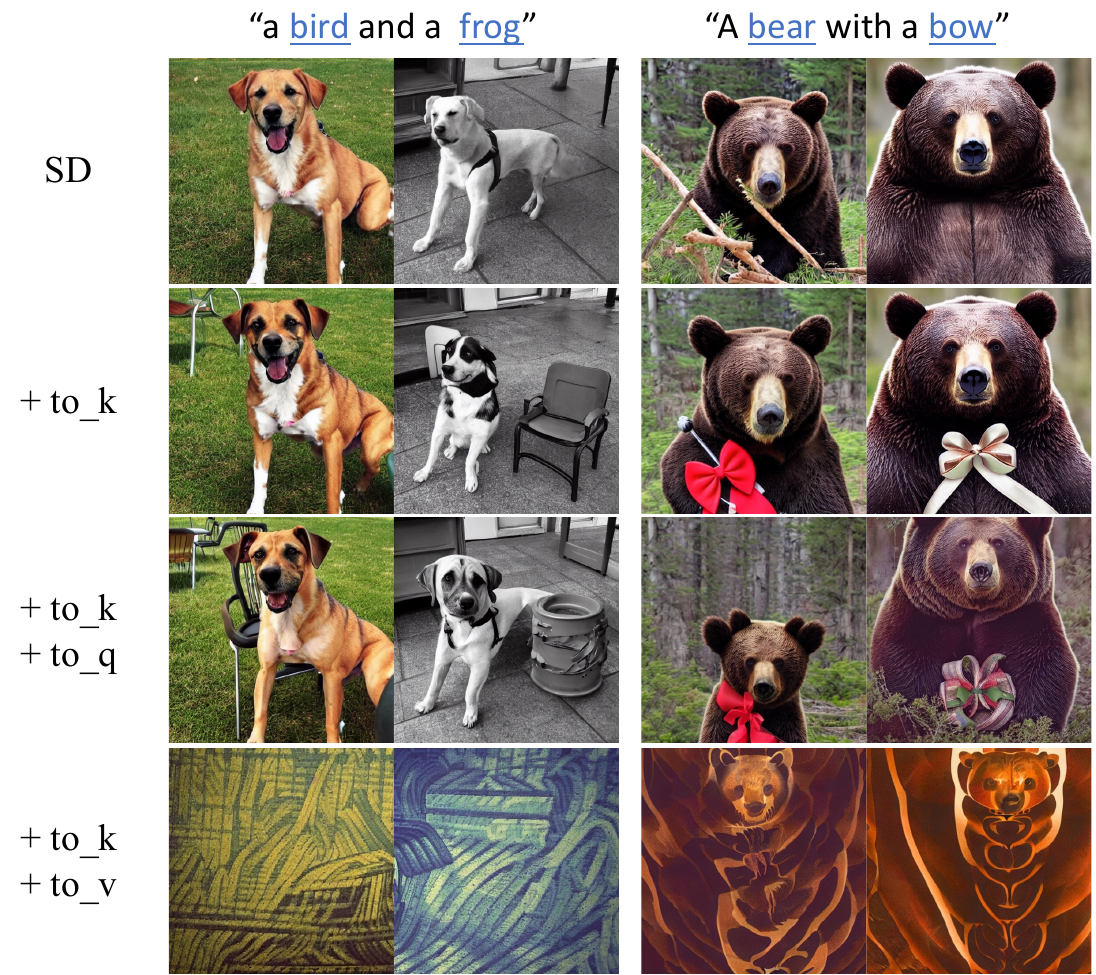}
    \vspace{-0.2in}
    \caption{Ablation study with tuning different sets of parameters. Tuning \emph{to\_v} functions leads to a decreased visual quality while tuning \emph{to\_q} functions has few differences. Our design of only finetuning the \emph{to\_k} functions leads to an effective and efficient solution.}
    \label{fig:qkv}
    \vspace{-0.1in}
\end{figure}

\begin{table}[t]
\centering
\resizebox{\linewidth}{!}{
\begin{tabular}{l|ccc} 
\toprule
 Method & FID ($\downarrow$) & Average Similarity Score ($\uparrow$) & Success Rate ($\uparrow$) \\
 \hline
 StableDiffusion & \color{gray} 32.96 &  0.742$\pm$0.091 & 0.209 \\ 
 + \emph{to\_q} & 39.71 & 0.781 $\pm$ 0.102 & 0.365 \\
 + \emph{to\_q} + \emph{to\_k} & 51.60 & 0.759 $\pm$ 0.099 & 0.347\\
 + \emph{to\_k} + \emph{to\_v}  & 445.01 & 0.213 $\pm$ 0.051 & 0.004\\ \hline
 + \emph{to\_k} (ours) & \bf 36.85 & \bf 0.809 $\pm$ 0.086 & \bf 0.410 \\
 \bottomrule
\end{tabular}
}
\caption{Quantitative comparison with finetuning different parameters. Updating \emph{to\_v} will hurt the performance while finetuning \emph{to\_q} has few contributions. Tweaking \emph{to\_k} is the optimal choice.  
}
\label{tab:qkv}
\vspace{-0.1in}
\end{table}

\smallsec{Fintuning different parameter groups.} In Section~\ref{sec:method}, we depot the key role of fintuning the key mapping (\emph{to\_k}) functions for cross-attention modules. To validate our design, we further build 3 additional variants that (1) finetunes both key mapping and value mapping functions (\emph{to\_v}); (2) finetunes both key mapping and query mapping (\emph{to\_q}) functions; and (3) only finetunes query mapping functions. Both Figure~\ref{fig:qkv} and Table~\ref{tab:qkv} show that tuning the value mapping functions will lead to a decreased performance, verifying the role of the value mapping function in forming the object representations for which we do not want to tune from Stable Diffusion. Moreover, the query mapping function plays a less important role in the compositional synthesis task, involving or ignoring it has few differences while our original model design without tuning the query mapping function is effective and efficient.

\begin{figure}[t]
    \centering
    \includegraphics[width = \linewidth]{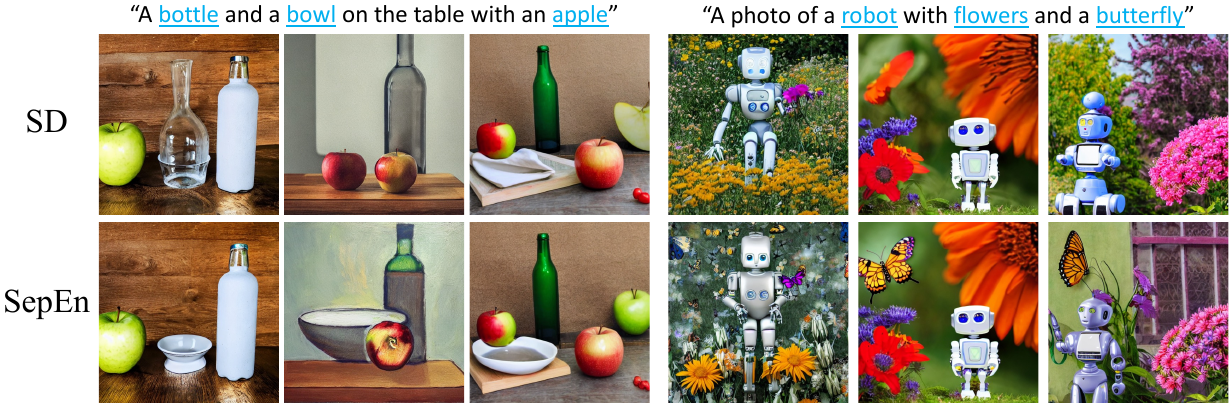}
    
    \caption{Extension of our method to more than two concepts. Our method is able to synthesize high-quality images matching the input prompts, indicating the good generalizability. }
    \label{fig:multi}
    \vspace{-0.1in}
\end{figure}

\smallsec{Extension to more than two concepts} In Figure~\ref{fig:multi}, we show the generalization of our finetuning algorithm to more than two concepts. Our model is still able to synthesize high-quality images matching the text prompts, indicating the decent generalizability of the proposed method. 

\section{Limitation and Conclusion}
\label{sec:conclusion}

\begin{figure}[t]
    \centering
    \includegraphics[width = \linewidth]{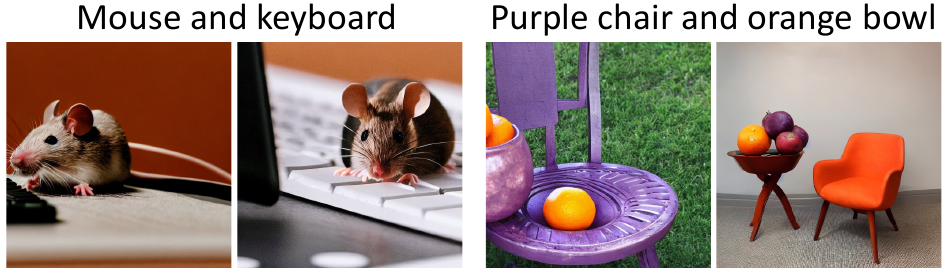}
    
    \caption{Failure case: after large-scale finetuning, the model fails to distinguish the polysemy words. The model fails to distinguish the digital mouse from the animal mouse, and the fruit orange from the color orange. A better language model such as LLaMA~\cite{touvron2023llama} and a more diverse training process can be helpful.}
    \label{fig:fail}
    \vspace{-0.1in}
\end{figure}

We observed one typical failure case of our method after large-scale finetuning: failing to distinguish the meaning of polysemy words. We show two typical examples in Figure~\ref{fig:fail}, where the model fails to distinguish the digital mouse from the animal mouse, and the fruit orange from the color orange. We think a better language model such as LLaMA~\cite{touvron2023llama} and a more diverse training process can help with this kind of failure case.

In summary, we addressed the limitations of diffusion-based T2I models in accurately generating images based on textual inputs, particularly its compositional misalignment issue. We propose the compositional finetuning strategy \textbf{Separate-and-Enhance} by incorporating two novel loss functions.
Through a complete evaluation experimental evaluation, we demonstrate that our method design is promising for enhancing the compositional capacity of diffusion models. We also showcase that the proposed method is generally helpful for a large collection of concepts and has a great generalization capacity.

{
    \small
    \bibliographystyle{ieeenat_fullname}
    \bibliography{main}
}

\clearpage 
\newpage
\appendix
\setcounter{figure}{0}
\setcounter{table}{0}
\renewcommand{\thefigure}{\Alph{figure}}
\renewcommand{\thetable}{\Alph{table}}


\section{Additional Ablations and Discussions}

\subsection{FID Evaluation for Large-Scale Experiments}

In the main paper, we only report the Average Similarity Score and Success Rate for the large-scale experiment. The reason is that FID is not applicable to the Stable Diffusion baseline, mainly due to its tendency to underperform in generating images with multiple objects, despite generating photorealistic single-object images. We additionally report the FID scores for the large-scale experiment in Table~\ref{suptab:fid} for reference. The source images used to compute the FID are generated by the pre-trained Stable Diffusion with the single concept with the format of ``A photo of a $<$concept$>$''. Since the ratio of unseen objects increases from seen-seen (0\%) to unseen-unseen (100\%) while they remain at a fixed ratio in the source images (9.09\%), the FID increases from seen-seen to unseen-unseen. Notably, our model even reaches a lower FID score in the most challenging unseen-unseen setting, indicating that our model maintains a good quality for the generated images after finetuning. Combining with Table~\ref{tab:large}, Figure~\ref{fig:cover} and \ref{fig:large_scale} in the main paper, we are able to generate significantly better-aligned images while maintaining good visual qualities.

\begin{table}[t]
\centering
\resizebox{\linewidth}{!}{
\begin{tabular}{l|ccc} 
\toprule
Model & Seen-Seen & Seen-Unseen & Unseen-Unseen \\ \hline
Stable Diffusion~\cite{rombach2022high} & \bf 20.07 & \bf 43.25 & 83.54 \\ 
+ \modelname & 26.01 & 45.04 & \bf 80.08 \\
\bottomrule
\end{tabular}
}
\caption{FID comparison for Stable Diffusion~\cite{rombach2022high} and our method under the large-scale setup. Combining with Table~\ref{tab:large} in the main paper, we are able to generate significantly better-aligned images while keeping good visual qualities.}
\label{suptab:fid}
\vspace{-0.2in}
\end{table}

\subsection{Single Prompt v.s. Large-Scale Training}

\begin{figure*}[t]
    \centering
    \includegraphics[width = 0.8 \linewidth]{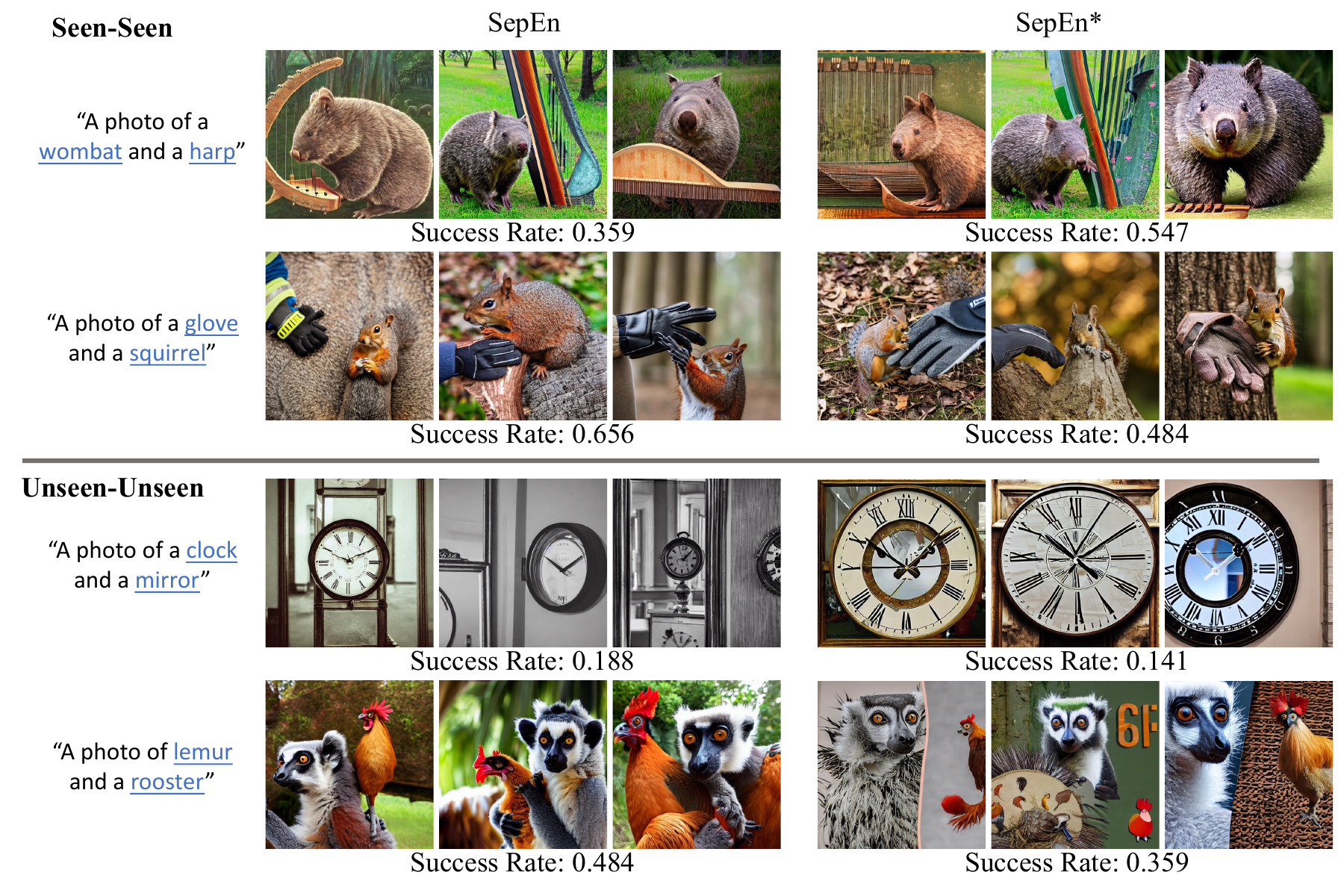}
    \caption{Visual comparisons between our method and the variant that individually optimize each prompt (SepEn$^*$). The compositional finetuning makes it possible for our method to scale up to a large collection of concepts to learn a global representation for multiple concepts with diffusion models, thereby outperforming the variant individually optimized for each single prompt. }
    \label{supfig:large}
\end{figure*}

To have a comprehensive understanding of our method, we additionally run two experiments for the large-scale setting under the seen-seen and unseen-unseen settings by adding a variant of our model, \modelnamestar, which does compositional finetuning on every single prompt. By comparing the results of \modelname and \modelnamestar, we aim to better understand the effect of our compositional finetuning regarding the scalability (seen-seen) and generalizability (unseen-unseen). The quantitative results are reported in Table~\ref{suptab:large} and a few qualitative results are included in Figure~\ref{supfig:large}. 

Surprisingly, we find that the large-scale model even outperforms the single-prompt models both qualitatively and quantitatively. We think the underlying factor is that by jointly training with a large collection of concepts, our method better models the relationships between multiple concepts, thereby being able to tackle the more challenging fine-grained concepts in the large-scale setup. These results also highlight the contribution of this paper -- the compositional finetuning makes it possible to scale up to a large collection of concepts to learn a global representation for multiple concepts with diffusion models.

\begin{table*}[t]
\centering
\resizebox{0.75 \linewidth}{!}{
\begin{tabular}{l|cc|cc} 
\toprule
\multirow{2}{*}
{\shortstack{Method}} & \multicolumn{2}{c|}{seen-seen}   & \multicolumn{2}{c}{unseen-unseen} \\ \cline{2-5}
    & Average Sim. Score ($\uparrow$) & Success Rate ($\uparrow$) & Average Sim. Score ($\uparrow$) & Success Rate ($\uparrow$) \\
 \hline
 StableDiffusion  & 0.641 $\pm 0.107$ &0.212& 0.633 $\pm$ 0.098 & 0.203\\ 
 + SepEn  & \bf 0.686 $\pm$ 0.107 & \bf 0.299 & 0.679 $\pm$ 0.102 & \bf 0.294 \\ \hline 
 + SepEn$^*$ & 0.681 $\pm$ 0.095 &  0.294 & \bf 0.687 $\pm$ 0.099 &  0.287 \\
 \bottomrule
\end{tabular}
}
\caption{Quantitative comparison between our method and the variant individually optimized for each prompt (SepEn$^*$). By joint training with a large collection of concepts, our method better models the relationships between multiple concepts, thereby being able to tackle the more challenging fine-grained concepts in the large-scale setup.}
\label{suptab:large}
\end{table*}

\subsection{Ablation on Learning Rate and Training Steps}

\begin{figure}[t]
    \centering
    \includegraphics[width = \linewidth]{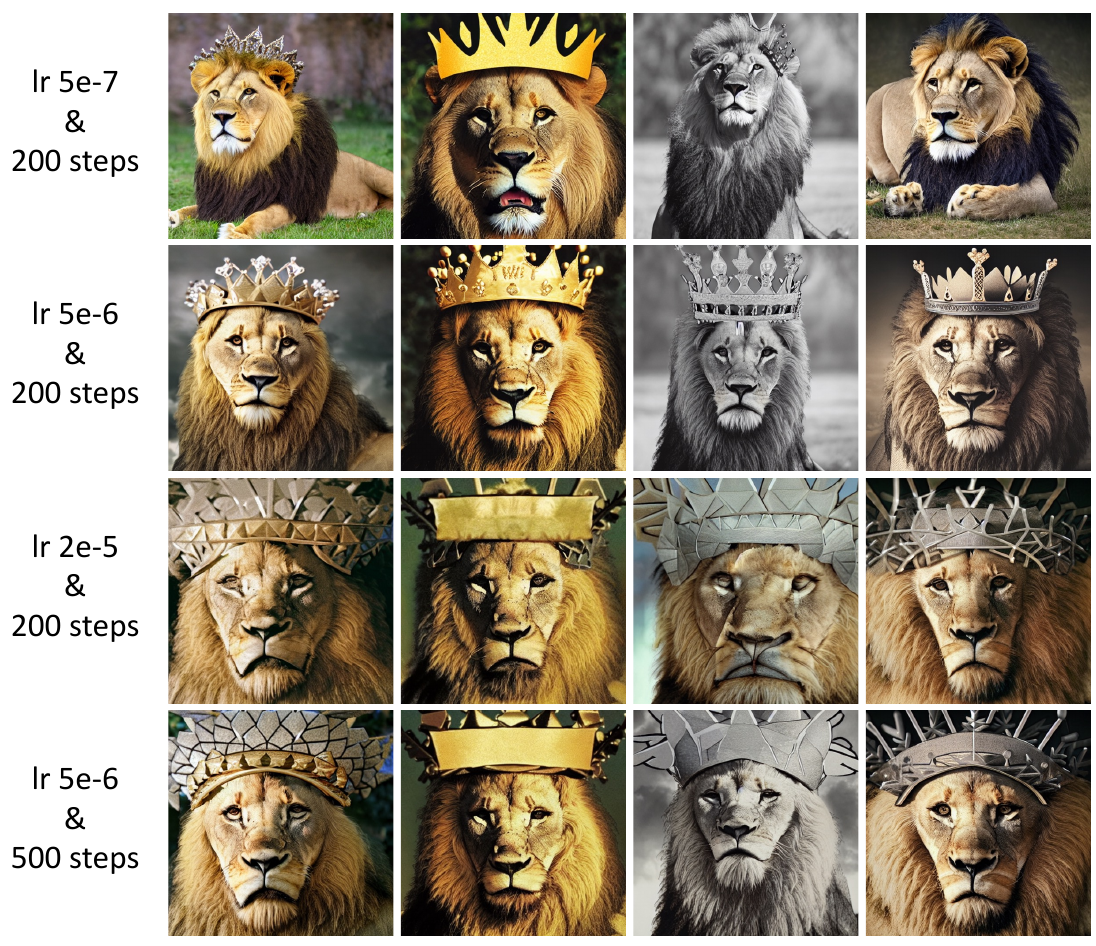}
    \caption{Visual comparisons for the variants of our model with differences in learning rates or training steps. A larger learning rate or longer training steps lead to distortion of the images with both worse visual quality and text-image alignment; while a smaller learning rate results in the underfitting of the model with undesired results. }
    \label{supfig:lr}
\end{figure}

We ablate the choice of the learning rate and training steps for single-concept experiments in this section. We add three variants with (1) a smaller learning rate \emph{5e-7}, (2) a larger learning rate \emph{2e-5}, and (3) a longer training step of \emph{500}. The quantitative results are shown in Table~\ref{suptab:lr} and qualitative comparisons for one prompt, ``A photo of a lion with a crown'' is shown in Figure~\ref{supfig:lr}. Based on these results, we find that a larger learning rate or longer training steps lead to distortion of the images with both worse visual quality and text-image alignment; while a smaller learning rate results in the underfitting of the model with undesired results. Our current learning choice of the learning rate and training steps yielded the best performance. 

\begin{table}[t]
\centering
\resizebox{\linewidth}{!}{
\begin{tabular}{cc|ccc} 
\toprule
Learning Rate & Steps & FID ($\downarrow$) & Avg. Similarity Score ($\uparrow$) & Success Rate ($\uparrow$)\\ \hline
2e-5 & 200 & 54.80 & 0.711 $\pm$ 0.105 & 0.388 \\
7e-7 & 200 & \bf 36.41 & 0.783 $\pm$ 0.083 & 0.372\\
 5e-6 & 500 & 51.94 & 0.724 $\pm$ 0.097 & 0.393\\ \hline
 5e-6 & 200 & 36.85 & \bf 0.809 $\pm$ 0.086 & \bf 0.410 \\
 \bottomrule
\end{tabular}
}
\caption{Ablation study on the learning rate and training steps for single-concept evaluations. A larger learning rate or longer training steps lead to distortion of the images with both worse visual quality and text-image alignment; while a smaller learning rate results in the underfitting of the model with undesired results. }
\label{suptab:lr}
\end{table}

\subsection{Inference time}

As stated in the main paper, one limitation of the test-time-adaptation-based methods~\cite{chefer2023attend,agarwal2023star} is the longer inference times owing to the adaptations performed during testing. We measure the average inference time by generating a single image for each single-concept test prompt evaluated in Section 4.1. The results for Stable Diffusion~\cite{rombach2022high}, Attend-and-Excite~\cite{chefer2023attend} and our \modelname are reported in Table~\ref{suptab:time}. Those results are measured based on one single NVIDIA-A100 GPU. 
By doing the compositional finetuning with the two proposed objectives, we do not increase the inference time while significantly improving the text-image alignment.

\begin{table}[t]
\centering
\resizebox{0.75 \linewidth}{!}{
\begin{tabular}{l|c} 
\toprule
Model & Average Inference Time (s) \\ \hline
Stable Diffusion~\cite{rombach2022high} & \bf 2.43 \\ 
+ A\&E ~\cite{chefer2023attend} & 11.69 \\
+ \modelname & 2.45 \\
\bottomrule
\end{tabular}
}
\caption{Average inference time comparison for our method and the baselines. By doing the compositional finetuning with the two proposed objectives, we do not increase the inference time while significantly improving the text-image alignment.}
\label{suptab:time}
\end{table}

\section{Experimental Evaluation Details}

\subsection{Baseline Details}

\smallsec{Stable Diffusion~\cite{rombach2022high}.} We use the official implementation from Diffuser\footnote{\url{https://github.com/huggingface/diffusers}}. We use the pretrained checkpoint from version v1.4. All the other baselines are built upon the same checkpoint. During run 50 denoising steps with a guidance scale~\cite{ho2022classifier} of 6.0 for inference. All the other models follow the same inference setup. 

\smallsec{Attend-and-Excite~\cite{chefer2023attend}.} We use the official open-source implementation\footnote{\url{https://github.com/yuval-alaluf/Attend-and-Excite}} of Attend and Excite. We use their default configurations for the single-prompt evaluation which runs the test-time-adaptation for the first 25 denoising steps. 

\smallsec{SepEn (TTA).} We build this variant based on the codebase of Attend and Excite. We use the same configurations with them but change their loss functions with ours. 

\smallsec{Other Baselines.} Due to a lack of public implementation of \textbf{A-Star}~\cite{agarwal2023star}, we report their original numbers. We compute the Average Similarity Score by running a weighted average strategy of the three sub-numbers they reported. We also cite the results of Composable Diffusion~\cite{liu2022compositional} and Structure Diffusion~\cite{feng2022training} for single-prompt evaluation from the A-Star paper.

\subsection{Evaluation Metrics Details and Discussions}

Notice that currently there are no best evaluation metrics to measure the compositional ability of the T2I models. Previous work~\cite{agarwal2023star,chefer2023attend} mainly uses the Average Similarity Score to measure the performance. However, this metric has its own limitations. Therefore, we further introduce two additional metrics to have a comprehensive evaluation. We discuss the details of the three metrics below.

\smallsec{FID} measures the realism of the generated images, by computing the Fr{\'e}chet distance between two Gaussians fitted to feature representations of the source images and the target images~\cite{dowson1982frechet,parmar2021cleanfid}. However, since there are few multi-object images for the test prompts in the real world, and SD is renowned for its capacity to generate high-quality images (which we aim to preserve after finetuning), we use generated images for \emph{single-object} prompts from SD as the source images. The FID scores can reflect the realism of the generated images to some extent while are also affected by the bias of the single-prompt images.
We use the public implementation\footnote{\url{https://github.com/GaParmar/clean-fid}} of the CleanFID~\cite{parmar2021cleanfid} for computing the FID scores.

\smallsec{Average Similarity Score} is proposed by Chefer \etal~\cite{chefer2023attend} to measure if the generated contents match the input prompts.
Following A\&E~\cite{chefer2023attend}, for each prompt, we compute the average BLIP~\cite{li2022blip} cosine similarity between the text prompt and the corresponding set of generated images. For the details of this metric, refer to Section A.4 from the supplementary of A\&E~\cite{chefer2023attend}. One limitation of this metric is the bias of the BLIP model: comparing the results of Table 1 and Table 2 in the main paper, the overall BLIP score is much lower for all the models under the large-scale setup. One reason is that we have a lot of fine-grained categories in the large-scale setup while the BLIP model may not be able to distinguish them. For instance, the category ``lemur'' is usually represented by ``monkey'' for the BLIP model. 

\smallsec{Success Rate} measures if the output images contain all objects mentioned in the text prompt. We use a detection model, Detic~\cite{zhou2022detecting}, pre-trained on ImageNet-21K to detect all possible target objects from the given prompt. Concretely, this detector runs object classification for each region proposal. We first filter out the low-confident bounding boxes with a threshold $\alpha$, then we count as a success case if all target object labels from the prompt are detected. $\alpha$ is set to 0.7 for single-prompt evaluation and 0.3 for large-scale experiments, as some of the concepts for large-scale training are very challenging for the detector. One limitation of this metric is that, we do not take the number of generated objects into consideration. For instance, if the model generates several balloons while the input prompt is specified as \emph{one} balloon, we still count it as a success case.

\subsection{Prompts for Large-Scale Experiments}

Following the single-prompt evaluation setup, which was originally proposed by Attend-and-Excite~\cite{chefer2023attend}, we focus on two types of concepts: animals and objects. We use ChatGPT~\cite{chatgpt} to generate 110 different animals and 110 different objects by querying it with \emph{Could you please generate the name for 110 different single-word animals (objects)?} Then we filter out repeated ones, and the ones not in the label set of ImageNet-21K~\cite{deng2009imagenet}. We query ChatGPT again with \emph{Could you generate X more?} until the number of animals/objects reaches 110. We held out 1 animal/object for every 11 as the unseen concept and the others as the training concepts. Notice that even for large-scale experiments, we \textbf{do not} require additional data for finetuning our model, but select two concepts each time to finetune \emph{only} based on the prompts. This strategy makes it possible for our method to easily generalize to large-scale training set-ups, making our method further distinguished from the previous test-time-adaptation-based methods~\cite{agarwal2023star,liu2022compositional,chefer2023attend}.

We conducted three different types of evaluations for the large-scale experiment. For \textbf{Seen-Seen} evaluation, we randomly form 100 pairs for the whole 200 concepts. The prompt format is ``A photo of a $<$seen\_concept\_1$>$ and a $<$seen\_concept\_2$>$''. For \textbf{Seen-Unseen} evaluation, we randomly select 4 concepts from the training set for each unseen concept, resulting in 80 prompts in total for Seen-unseen evaluation. The prompt format is ``A photo of a $<$unseen\_concept\_1$>$ and a $<$seen\_concept\_1$>$''. For \textbf{Unseen-Unseen} evaluation, for each concept in the unseen set, we randomly select another concept from the unseen set. The prompt format is ``A photo of a $<$unseen\_concept\_1$>$ and a $<$unseen\_concept\_2$>$''. We include these 200 prompts we used in the supplementary zip file (large\_scale\_prompts.txt).

\subsection{Implementation Details}

We build our model based on Stable Diffusion v1.4. We set $\lambda_E$ to 1.0 and $\lambda_E$ to 2.0 based on experience and attempts. We set $\lambda_N$ to 0.0 for individual training and 0.5 for large-scale training. We tune our method for 200 steps for each pair of individual concepts with a batch size of 4 on a single NVIDIA A-100 GPU. For the large-scale experiments, we tuned our model for 10,000. We use Adam~\cite{kingma2014adam} for optimization for all our experiments. For single-prompt evaluation, we use a fixed learning rate of $5e-6$; while for the large-scale experiment, we use a cosine annealing scheduler with a warm-up step of 300, and a peak learning rate of $1e-6$. For single prompt evaluation, it takes around 2 minutes to run our model and around 4 hours for the large-scale experiment with a single NVIDIA A-100 GPU.

\section{More Visualizations}

\begin{figure*}[t]
    \centering
    \includegraphics[width =  \linewidth]{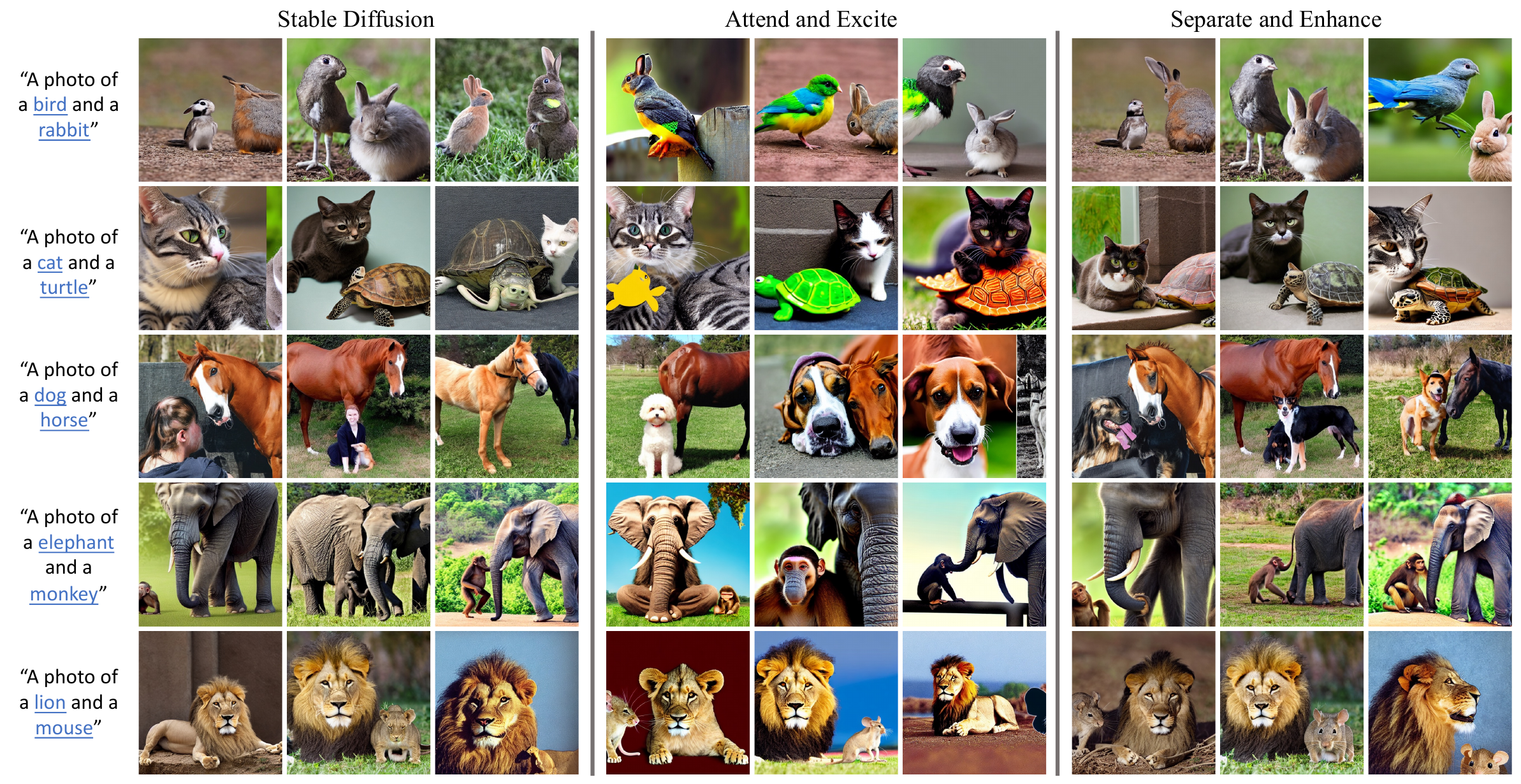}
    \caption{Visual comparisons for the single-prompt evaluations for animal-animal concept pairs. Our method outperforms the baselines under both realism and text-image alignment, owing to the two proposed novel objectives.}
    \label{supfig:animal-animal}
\end{figure*}

\begin{figure*}[t]
    \centering
    \includegraphics[width = \linewidth]{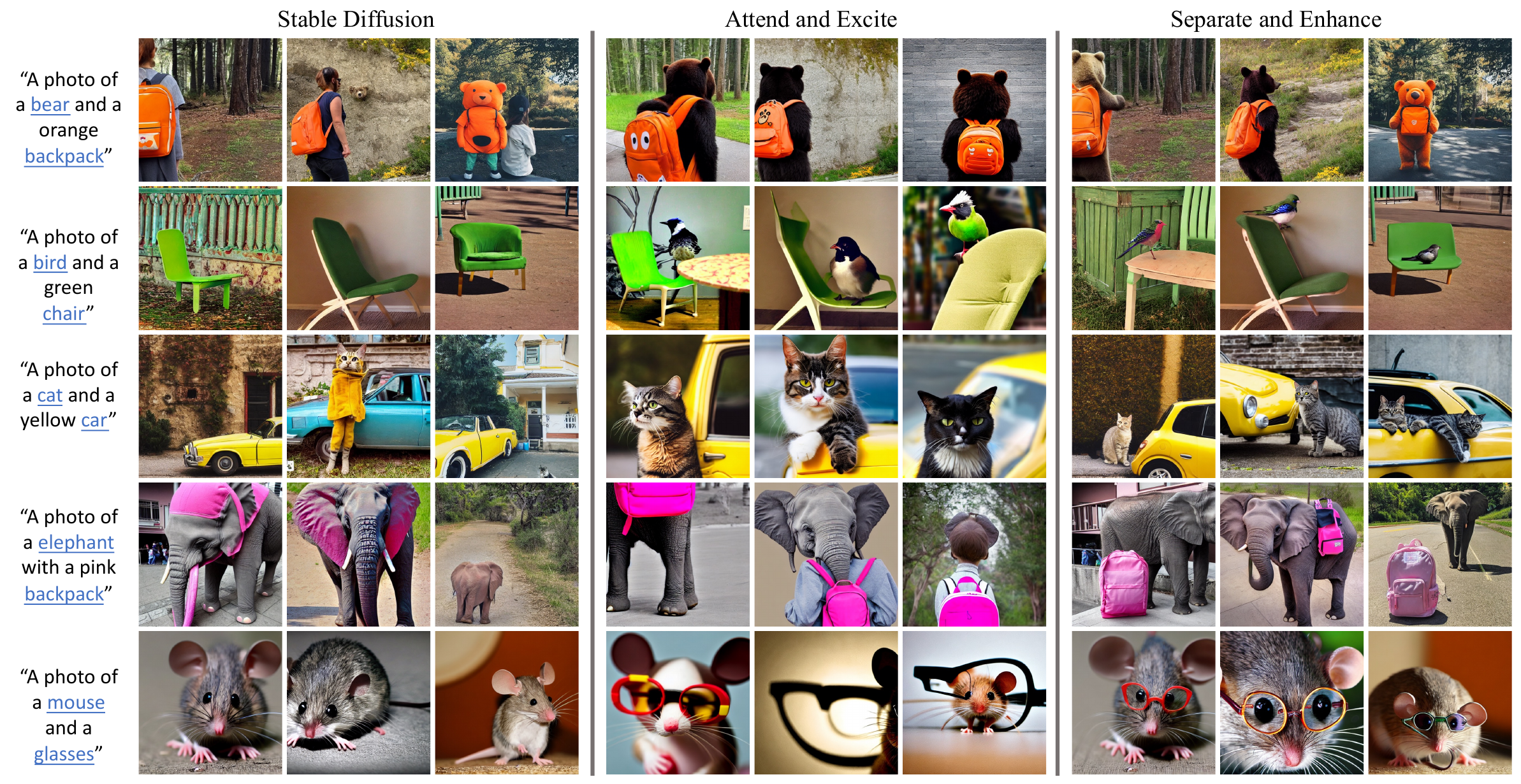}
    \caption{Visual comparisons for the single-prompt evaluations for animal-object concept pairs. Our method outperforms the baselines under both realism and text-image alignment, owing to the two proposed novel objectives.}
    \label{supfig:animal-object}
\end{figure*}

\begin{figure*}[t]
    \centering
    \includegraphics[width = \linewidth]{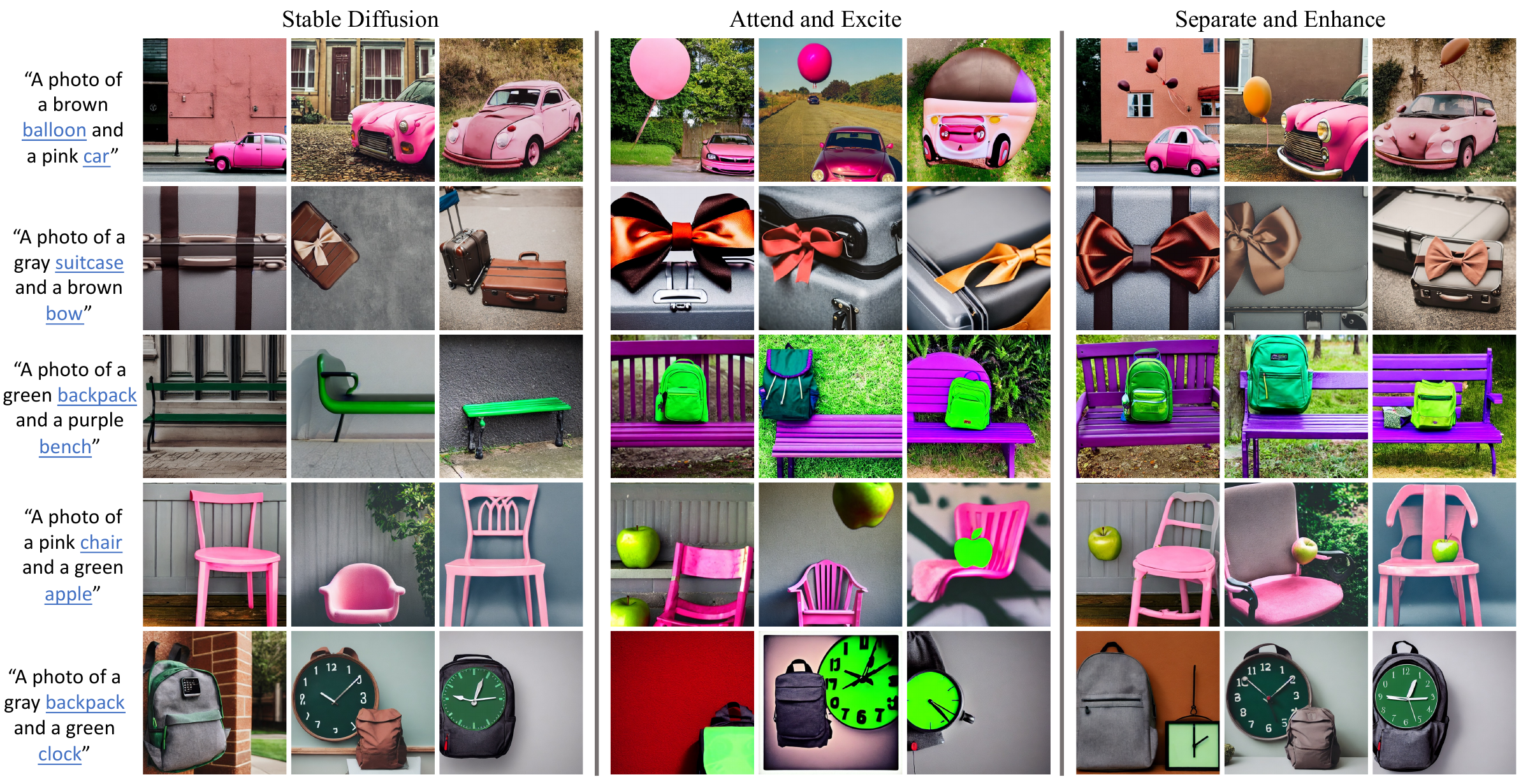}
    \caption{Visual comparisons for the single-prompt evaluations for object-object concept pairs. Our method outperforms the baselines under both realism and text-image alignment, owing to the two proposed novel objectives.}
    \label{supfig:object-object}
\end{figure*}

\begin{figure*}[t]
    \centering
    \includegraphics[width = 0.75 \linewidth]{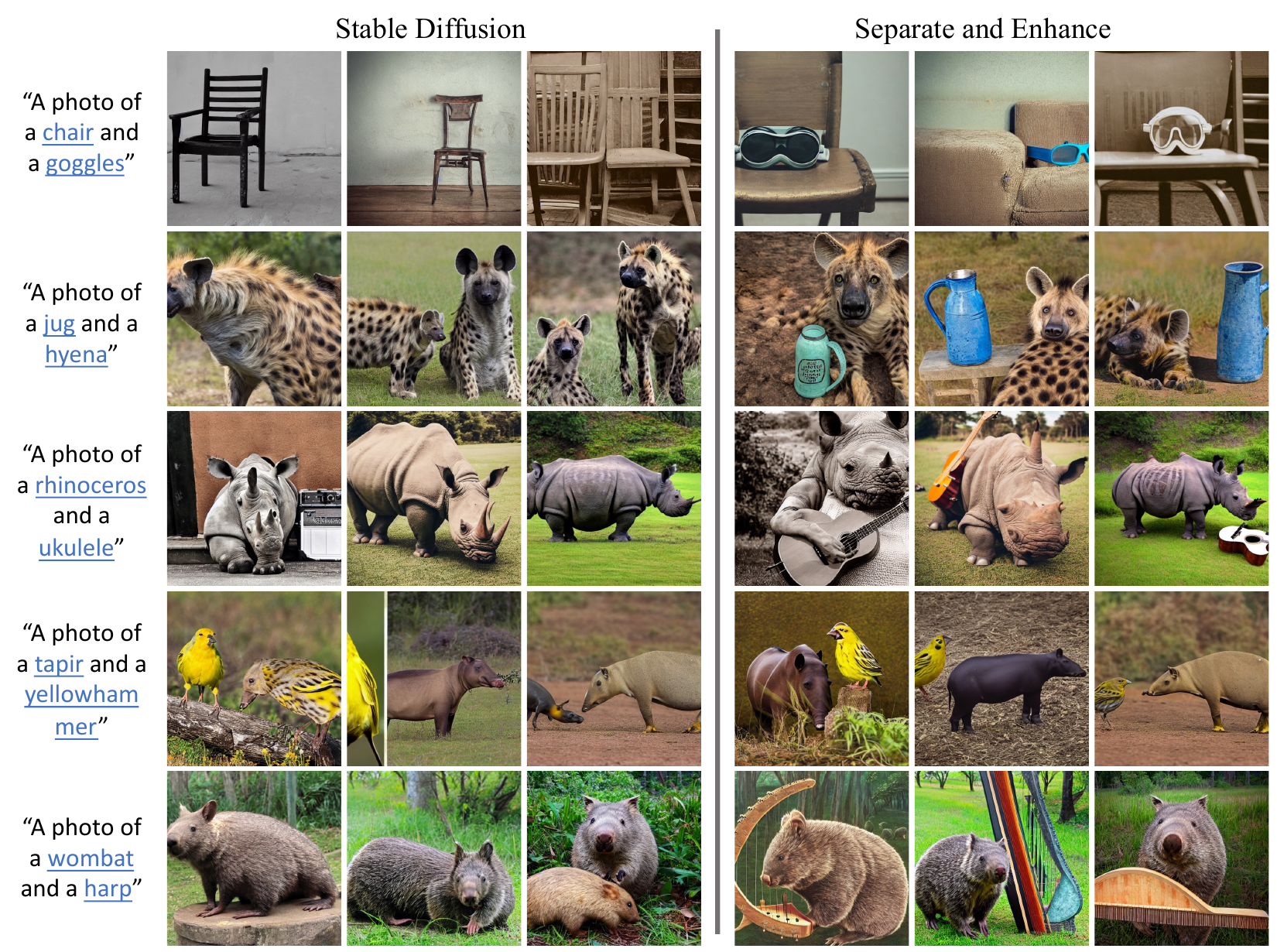}
    \caption{Visual comparisons for the large-scale experiment under the seen-seen setting. our method has better text-image alignments compared with the baseline Stable Diffusion~\cite{rombach2022high} while maintaining a good visual quality.}
    \label{supfig:large-ss}
\end{figure*}

\begin{figure*}[t]
    \centering
    \includegraphics[width = 0.75 \linewidth]{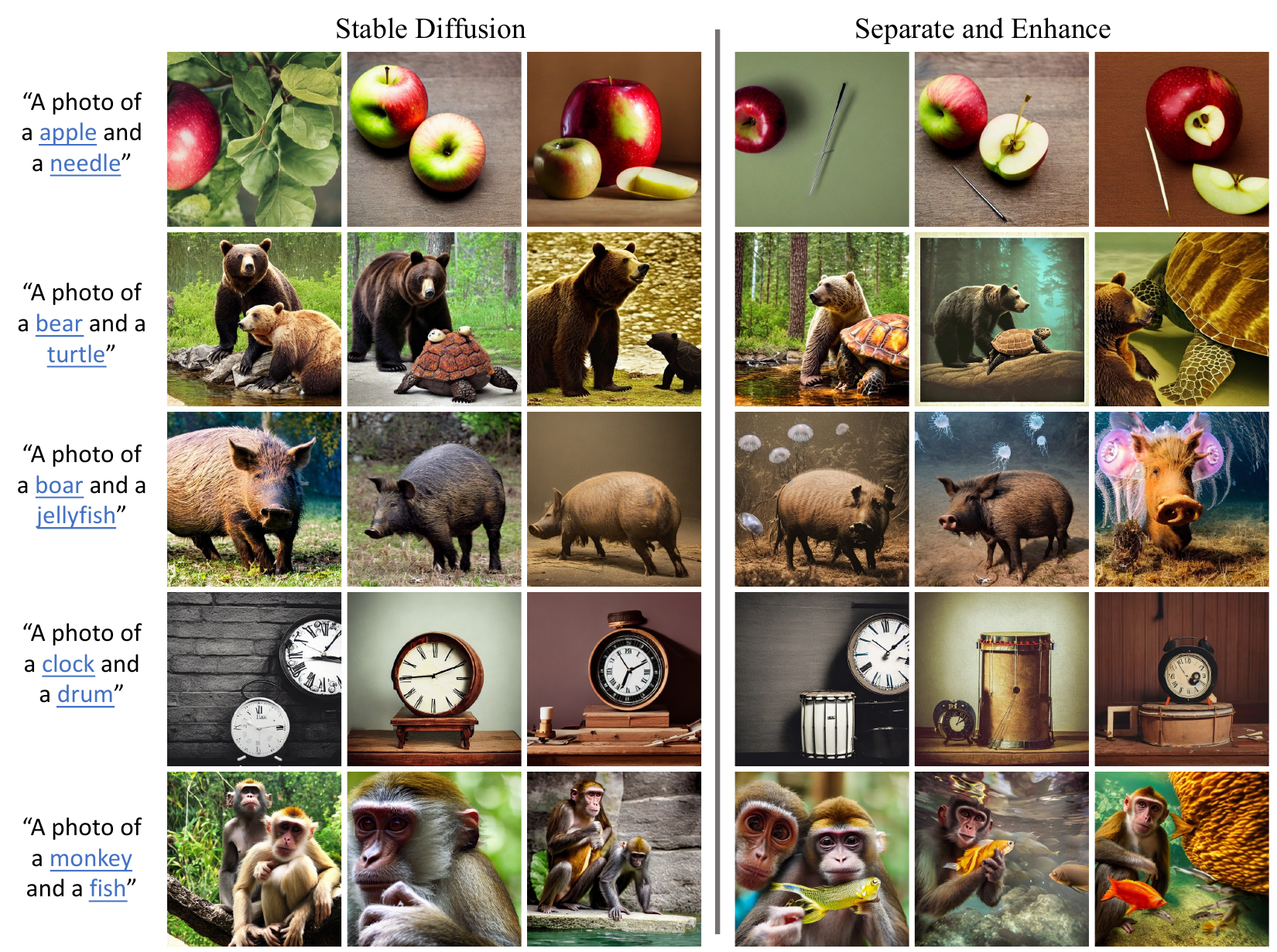}
    \caption{Visual comparisons for the large-scale experiment under the seen-unseen setting. our method has better text-image alignments compared with the baseline Stable Diffusion~\cite{rombach2022high} while maintaining a good visual quality.}
    \label{supfig:large-su}
\end{figure*}

\begin{figure*}[t]
    \centering
    \includegraphics[width = 0.75 \linewidth]{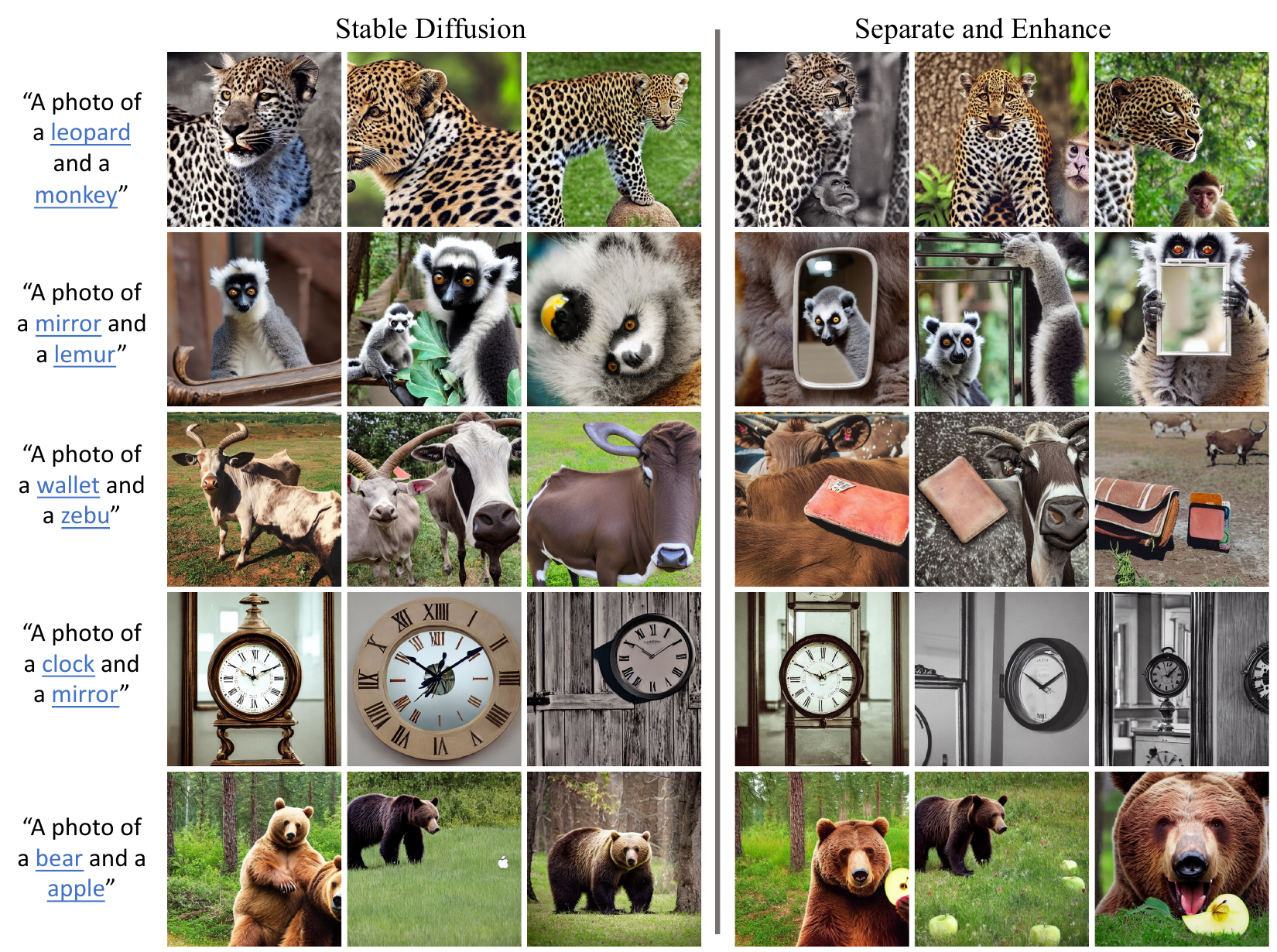}
    \caption{Visual comparisons for the large-scale experiment under the unseen-unseen setting. our method has better text-image alignments compared with the baseline Stable Diffusion~\cite{rombach2022high} while maintaining a good visual quality.}
    \label{supfig:large-uu}
\end{figure*}

For single-prompt evaluations, we show the comparison among Stable Diffusion~\cite{rombach2022high}, Attend-and-Excite~\cite{chefer2023attend} and ours for ``animal-animal'', ``animal-object'', and ``object-object'' evaluations in Figure~\ref{supfig:animal-animal}, \ref{supfig:animal-object}, and \ref{supfig:object-object} respectively. We also show the visual comparisons between Stable Diffusion and ours under the large-scale setup in Figure~\ref{supfig:large-ss} (seen-seen), Figure~\ref{supfig:large-su} (seen-unseen), and Figure~\ref{supfig:large-uu} (unseen-unseen). Our method outperforms the baselines under both realism and text-image alignment, owing to the two proposed novel objectives.

\end{document}